\documentclass[nojss]{jss} 

\usepackage{orcidlink,thumbpdf,lmodern}

\usepackage{booktabs, multirow, adjustbox}  
\usepackage{framed}
\usepackage{amsmath}
\usepackage{amsfonts}
\usepackage{amssymb}
\usepackage{mathtools}
\usepackage{enumitem}  
\usepackage{caption}
\usepackage{subcaption}
\usepackage{algorithm}
\usepackage{algpseudocode}

\newcommand{\x}{\boldsymbol{x}}
\newcommand{\z}{\boldsymbol{z}}
\newcommand{\w}{\boldsymbol{w}}
\newcommand{\W}{\boldsymbol{W}}
\newcommand{\xsb}{{\boldsymbol{x}_{\thickbar{\mathcal{S}}}}}
\newcommand{\Xsb}{{\boldsymbol{X}_{\thickbar{\mathcal{S}}}}}
\newcommand{\Xs}{{\boldsymbol{X}_{\mathcal{S}}}}
\newcommand{\xs}{{\boldsymbol{x}_{\mathcal{S}}}}
\newcommand{\xss}{{\boldsymbol{x}_{\mathcal{S}}^*}}
\newcommand{\sbb}{{\thickbar{\mathcal{S}}}}
\newcommand{\s}{{\mathcal{S}}}
\newcommand{\M}{{\mathcal{M}}}
\newcommand{\D}{{\mathcal{D}}}

\newcommand{\R}{\mathbb{R}}  
\renewcommand{\E}{\mathbb{E}_{\boldsymbol{x}_{\thickbar{\mathcal{S}}}}}  
\newcommand{\bphi}{{\boldsymbol{\phi}}}
\newcommand{\bmu}{{\boldsymbol{\mu}}}

\newcommand{\bSigma}{{\boldsymbol{\Sigma}}}
\newcommand{\N}{\mathbb{N}}   

\newcommand{\independence}{{\code{independence}}}

\newcommand{\empirical}{{\code{empirical}}}

\newcommand{\gaussian}{{\code{gaussian}}}

\newcommand{\ctree}{{\code{ctree}}}

\newcommand{\vaeac}{{\code{vaeac}}}

\usepackage{accents}
\newcommand{\thickbar}{} 
\DeclareRobustCommand*\thickbar[1]{\accentset{\rule{.35em}{.65pt}}{#1}}

\newcommand{\diff}{\mathop{}\!\mathrm{d}}

\DeclareMathOperator{\doop}{do}
\DeclareMathOperator*{\argmin}{arg\,min}
\newcommand{\set}[1]{\left\lbrace #1 \right\rbrace}

\author{Martin Jullum~\orcidlink{0000-0003-3908-5155}\\Norwegian Computing Center, Norway
   \And Lars Henry Berge Olsen~\orcidlink{0009-0006-9360-6993}\\University of Oslo, Norway\\Norwegian Computing Center, Norway
   \AND Jon Lachmann~\orcidlink{0000-0001-8396-5673}\\Indicio Technologies, Sweden
   \And Annabelle Redelmeier\\Norwegian Computing Center, Norway
   }
\Plainauthor{Martin Jullum, Lars Henry Berge Olsen, Jon Lachmann, Annabelle Redelmeier}

\title{\pkg{shapr}: Explaining Machine Learning Models with Conditional Shapley Values in \proglang{R} and \proglang{Python}}
\Plaintitle{shapr: Explaining Machine Learning Models with Conditional Shapley Values in R and Python}
\Shorttitle{\pkg{shapr}: Conditional Shapley Value Explanations}

\Abstract{
This paper introduces the \pkg{shapr} \proglang{R} package, a versatile tool for generating Shapley value-based prediction explanations for machine learning and statistical regression models. 
Moreover, the \pkg{shaprpy} \proglang{Python} library brings the core capabilities of \pkg{shapr} to the \proglang{Python} ecosystem.
Shapley values originate from cooperative game theory in the 1950s, but have over the past few years become a widely used method for quantifying how a model's features/covariates contribute to specific prediction outcomes.
The \pkg{shapr} package emphasizes \textit{conditional} Shapley value estimates, providing a comprehensive range of approaches for accurately capturing feature dependencies -- a crucial aspect for correct model explanation, typically lacking in similar software. 
In addition to regular tabular data, the \pkg{shapr} \proglang{R} package includes specialized functionality for explaining time series forecasts. 
The package offers a minimal set of user functions with sensible default values for most use cases while providing extensive flexibility for advanced users to fine-tune computations. 
Additional features include parallelized computations, iterative estimation with convergence detection, and rich visualization tools. 
\pkg{shapr} also extends its functionality to compute \textit{causal} and \textit{asymmetric} Shapley values when causal information is available. 
Overall, the \pkg{shapr} and \pkg{shaprpy} packages aim to enhance the interpretability of predictive models within a powerful and user-friendly framework.
}

\Keywords{Explainable artificial intelligence, XAI, interpretable machine learning, IML, prediction explanation, Shapley values, feature dependence}
\Plainkeywords{Explainable artificial intelligence, XAI, interpretable machine learning, IML, prediction explanation, Shapley values, feature dependence}

\Address{
  Martin Jullum\\
  Norwegian Computing Center\\
  Gaustadalleen 23a, 4th floor\\
  Kristen Nygaards hus\\
  0373 Oslo\\
  Norway \\
  E-mail: \email{jullum@nr.no}
}

\begin{document}

\newpage
\section{Introduction} \label{sec:intro}

Understanding how complex predictive models produce their outcomes is crucial for their practical application, particularly in high-stakes contexts where trust, transparency, and accountability are essential.
The inherent trade-off between model complexity and interpretability often leaves simpler, more interpretable models behind, favoring advanced statistical regression and machine learning models such as generalized additive models (with higher-order interactions), support vector machines, (tree-based) boosting and bagging models, neural networks, and others. 
As a result, the growing demand for understanding how these high-performance models operate has led to a surge of research in the fields of eXplainable AI (XAI) and Interpretable Machine Learning (IML).
During the past few years, the Shapley value framework has established itself as the most prominent framework in this domain.


The Shapley value \citep{shapley1953value} originates from cooperative game theory, where it is used to distribute the payoff of a cooperative game to the players based on their contribution. 
In the context of XAI/IML, the framework is extensively used as a \textit{model-agnostic local} explanation framework to explain a prediction $f(\x^*)$ from a predictive model $f(\cdot)$. \textit{Model-agnostic} means it can explain any predictive model $f$, and local means it explains the prediction of a single, specific set of feature (covariate) values $\x^*$.
The Shapley value of feature $j$ is given by the formula
\begin{align} 
\phi_j = \sum_{\s \subseteq \M \backslash \{j\}} \frac{|\mathcal{S}|!(M-|\mathcal{S}|-1)!}{M!}\left(v(\mathcal{S} \cup \{j\}) - v(\s) \right), \label{eq:Shapley_value_formula}
\end{align}
where $\mathcal{M}=\{1, 2,\ldots,M\}$ is the set of the $M$ features, and $v(\mathcal{S})$ is the so-called characteristic/value/contribution function, which is some function representing the prediction with only the features in subset/coalition $\mathcal{S}$ present in the model\footnote{Although $v(\mathcal{S})$ also depends on the specific observation $\x^*$ being explained, we omit this dependence for notational convenience.}.
That is, the Shapley value measures how much a feature contributes to the prediction, averaged over all possible combinations of whether the other feature values are known or not.
In general, the Shapley values $\boldsymbol{\phi} = (\phi_0,\phi_1,\ldots,\phi_M)$ satisfies a series of beneficial properties such as summing to $v(\mathcal{S})-\phi_0$, where $\phi_0 = v(\emptyset)$, and $\phi_j$ can be roughly interpreted as the increase or decrease in the prediction caused by the knowledge of $x_j=x_j^*$, see e.g., \citet{shapley1953value, aas2019explaining} for details.

Using the Shapley value framework to explain model predictions was first proposed by \citet{strumbelj2010efficient}. However, the method did not gain widespread recognition until the publication of \citet{lundberg2017unified}, who defined
\begin{align}
    \label{eq:ContributionFunc}
    \begin{split}
        v(\mathcal{S}) 
        =
        \E\left[ f(\boldsymbol{x}) | \boldsymbol{x}_{\mathcal{S}} = \boldsymbol{x}_{\mathcal{S}}^* \right] 
        =
        \E\left[ f(\boldsymbol{x}_{\thickbar{\mathcal{S}}}, \boldsymbol{x}_{\mathcal{S}}) | \boldsymbol{x}_{\mathcal{S}} = \boldsymbol{x}_{\mathcal{S}}^* \right] 
        = 
        \int f(\boldsymbol{x}_{\thickbar{\mathcal{S}}}, \boldsymbol{x}_{\mathcal{S}}^*) p(\boldsymbol{x}_{\thickbar{\mathcal{S}}} | \boldsymbol{x}_{\mathcal{S}} = \boldsymbol{x}_{\mathcal{S}}^*) \diff \boldsymbol{x}_{\thickbar{\mathcal{S}}}.
    \end{split}
\end{align}
Here $\boldsymbol{x}_{\mathcal{S}} = \{x_j:j \in \mathcal{S}\}$ denotes the features in subset/coalition $\mathcal{S}$, $\boldsymbol{x}_{\thickbar{\mathcal{S}}} = \{x_j:j \in \thickbar{\mathcal{S}}\}$ denotes the features not in $\mathcal{S}$ (i.e., $\thickbar{\mathcal{S}} = \mathcal{M}\backslash\mathcal{S}$), $\E$ denotes the expectation over $\boldsymbol{x}_{\thickbar{\mathcal{S}}}$, and $p(\boldsymbol{x}_{\thickbar{\mathcal{S}}} | \boldsymbol{x}_{\mathcal{S}} = \boldsymbol{x}_{\mathcal{S}}^*)$ is the conditional density of $\xsb$ given $\xs = \xss$.

Although \citet{lundberg2017unified} defined $v(\mathcal{S})$ as a conditional expectation, their suggested estimation method actually estimates the following contribution function:
\begin{align}
    \label{eq:ContributionFuncMarg}
        v_{\text{marg}}(\mathcal{S}) 
        =
        \E\left[ f(\boldsymbol{x}_{\thickbar{\mathcal{S}}},\boldsymbol{x}^*_{\mathcal{S}}) \right] 
        = 
        \int f(\boldsymbol{x}_{\thickbar{\mathcal{S}}}, \boldsymbol{x}_{\mathcal{S}}^*) p(\boldsymbol{x}_{\thickbar{\mathcal{S}}}) \diff \boldsymbol{x}_{\thickbar{\mathcal{S}}}.
\end{align}
That is, the contribution function in \eqref{eq:ContributionFuncMarg} implicitly ignores the dependence between the features in $\boldsymbol{x}_{{\mathcal{S}}}$ and 
$\boldsymbol{x}_{\thickbar{\mathcal{S}}}$. This has significant consequences for the properties of the obtained Shapley values and is known to provide misleading conclusions in the presence of highly dependent features. 
This arises in part because \eqref{eq:ContributionFuncMarg} involves model evaluations at implausible feature combinations not seen during training, where the model’s output holds little practical relevance.
\citet{aas2019explaining} were the first to identify this issue and provide methods for properly providing Shapley values with the conditional expectation in \eqref{eq:ContributionFunc}. 
Despite the drawbacks, the benefits of significantly simpler computation 
and the availability of easy-to-use software (\pkg{shap}, \citet{shap}, in \proglang{Python}), have led to \eqref{eq:ContributionFuncMarg} still being very much in use \citep{chen2020true}. Shapley values computed with \eqref{eq:ContributionFuncMarg} are now often referred to as \textit{marginal} Shapley values, while those computed using \eqref{eq:ContributionFunc} are referred to as \textit{conditional} Shapley values. 

As mentioned above, explaining predictions with Shapley values helps quantify how valuable the observation of each feature is to a specific prediction. \textit{Conditional} Shapley values do this while respecting the distribution of the features, leading to explanations that are more relevant and realistic in practical use. 
As a result, using conditional Shapley values to explain predictions enables practitioners to extract relevant knowledge about the local behavior of black-box models.

The growing popularity of the Shapley value framework for prediction explanation has led to the emergence of numerous software packages in recent years.
The most popular software is the \pkg{shap} \citep{shap} library in \proglang{Python}, which implements a range of methods for computing/estimating Shapley values for different models and data types, including the model-agnostic KernelSHAP \citep{lundberg2017unified} and permutation-based formulation of \citet{strumbelj2010efficient, vstrumbelj2014explaining} (hereafter referred to as PermSHAP), and model-specific methods (TreeSHAP \citep{lundberg2020local2global} for tree-based models, and DeepLIFT for approximated Shapley values \citep{shrikumar2017learning} for neural networks). 
Building on the \code{PyTorch} framework, the \pkg{captum} \proglang{Python} library \citep{kokhlikyan2020captum} implements several explanation methods, including KernelSHAP and PermSHAP, as well as model-specific estimators.
The recent \pkg{shapiq} \citep{muschalik2024shapiq} \proglang{Python} library implements a wide range of methods for computing Shapley values for different types of games, including KernelSHAP and PermSHAP.

The \textit{DrWhy} universe is a collection of \proglang{R} packages for creating explanations and visual explorations of predictive models. It contains the \pkg{DALEX} \citep{DALEX_R} and \pkg{modelStudio} \citep{modelStudio} libraries, which work as high-level explanation tools where Shapley value-based prediction explanations are one of several ingredients. The actual Shapley value computations happen through sister packages: 
\pkg{shapper} \citep{shapper} is just a wrapper for the KernelSHAP implementation in the \pkg{shap} \proglang{Python} library \citep{shap}, 
\pkg{treeshap} \citep{treeshap} implements variants of the model-specific TreeSHAP  algorithm \citep{lundberg2020local2global}, 
\pkg{kernelshap} \citep{kernelshap} allows computing Shapley values through either KernelSHAP or PermSHAP, 
\pkg{fastshap} \citep{fastshap} also offers a fast implementation of the latter,  
while \pkg{iBreakDown} \citep{iBreakDown} uses so-called ``Break Down'' Tables \citep[Ch.~6]{ExplanatoryModelAnalysis} to approximate the Shapley values. 
The associated \pkg{shapviz} \citep{shapviz} package provides Shapley value visualizations.
Outside of the \textit{DrWhy} universe, the \pkg{iml} package \citep{iml} provides several interpretation/explanation methods. The package includes a method for deriving Shapley values for individual predictions based on PermSHAP. 

What is common for almost all of the above software packages is that they exclusively estimate, compute, or approximate \textit{marginal} Shapley values. While the so-called path-dependent variant of the TreeSHAP algorithm \citep{lundberg2020local2global} aims at estimating \textit{conditional} Shapley values, it is often (severely) biased in practice, see e.g., \citet[Sec.~4]{aas2019explaining} and \citet[Sec.~5.2]{chen2023algorithms}. 
The only library which touches upon \textit{conditional} Shapley values is the \pkg{shapiq} package, which, according to \citet[Appendix C]{muschalik2024shapiq}, offers a simple tree-based regression method for estimating the $v(\mathcal{S})$ in \eqref{eq:ContributionFunc}.

The \pkg{shapr} \proglang{R} package introduced in this paper implements an extended version of the KernelSHAP method \citep{Olsen2024improving} for approximating Shapley values, heavily focused on \textit{conditional} Shapley values. 
The core idea of the package is to be completely model-agnostic and offer a wide range of approaches for estimating $v(\mathcal{S})$ in \eqref{eq:ContributionFunc} (see Section \ref{sec:background:approaches}), allowing accurately estimated \textit{conditional} Shapley values to be computed for different types of features, dependencies, and distributions. 
Evaluation metrics for comparing different approaches are also readily available within the package. 
Combined with parallelized computations, convergence detection, progress updates, and extensive plotting functionality, the \pkg{shapr} package offers an efficient and user-friendly solution for estimating \textit{conditional} Shapley values. These accurate estimates are essential for an increased understanding of how features \textit{actually} contribute to model predictions in practice.
To increase the accessibility of the methodology, the \pkg{shapr} \proglang{R} package also comes with an accompanying \proglang{Python} wrapper called \pkg{shaprpy}. The wrapper makes it possible to explain models available in \proglang{Python} with the same estimation approaches and interface as the \proglang{R} package.

The present paper is based on \pkg{shapr} version 1.0.8 and \pkg{shaprpy} version 0.4.3. 
Note that \citet{Sellereite2019} briefly describes version 0.1.1 of \pkg{shapr}, which had significantly less functionality and used a different syntax compared to the current version\footnote{Version 1.0.0 of \pkg{shapr} represented a complete rewrite of the package, adding the majority of the functionality and flexibility in the current version, and enabling the creation of the \pkg{shaprpy} \proglang{Python} wrapper.}.

The rest of the paper is organized as follows: Section \ref{sec:background} provides descriptions of the main methodology implemented in \pkg{shapr}. 
In particular, it briefly describes all approaches used to estimate the $v(\mathcal{S})$ in \eqref{eq:ContributionFunc}. 
Section \ref{shapr:R} introduces the \pkg{shapr} package and its main functionality, and provides basic usage examples for estimating conditional Shapley value explanations. 
In Section \ref{sec:asymmetric_and_causal_shapley_values}, we introduce asymmetric and causal Shapley values, and show how such types of Shapley values can be computed with \pkg{shapr}, when causal information is available.
The associated \pkg{shaprpy} \proglang{Python} wrapper is introduced in Section \ref{sec:python}.
Section \ref{sec:conditional_shapley_values_forecast} describes functionality for explaining time series models with multiple forecasting horizons.
In Section \ref{sec:discussion}, we provide a summary and discuss potential future work.

\section{Methodological background} \label{sec:background}

Computing conditional Shapley values for prediction explanations involves two key steps: Obtaining accurate estimates of $v(\mathcal{S})$ in \eqref{eq:ContributionFunc}, and computing the Shapley values based on these estimates. 
Below, we briefly introduce the methodology implemented in the \pkg{shapr} package to address both steps.

\subsection{KernelSHAP} \label{sec:background:kernelshap}

The Shapley value formula in \eqref{eq:Shapley_value_formula} can be computationally intensive in high-dimensional settings, as its complexity grows exponentially with the number of features.
\citet{charnes1988extremal,lundberg2017unified} showed that the Shapley value formula in \eqref{eq:Shapley_value_formula} may also be conveniently expressed as the solution of the following weighted least squares (WLS) problem: 
\begin{align}
    \label{eq:ShapleyValuesDefWLS}
     \argmin_{\bphi \in \R^{M+1}}\sum_{\s \subseteq \M} k(M, |\s|)\Big(\phi_0 + \sum_{j \in \s}\phi_j - v(\s)\Big)^2,    
\end{align}
where 
\begin{align}
\label{eq:ShapleyKernelWeights}
    k(M, |\s|) = \frac{M-1}{\binom{M}{|\s|}|\s|(M-|\s|)}, 
\end{align}
for $|\s| = 0,1,2,\dots,M$, are the \textit{Shapley kernel weights} \citep{charnes1988extremal,lundberg2017unified}. In practice, the infinite Shapley kernel weights $k(M,0) = k(M, M) = \infty$ can be set to a large constant like $C = 10^6$ \citep{aas2019explaining}. 
The matrix solution of \eqref{eq:ShapleyValuesDefWLS} is 
\begin{align}
    \label{eq:ShapleyValuesDefWLSSolution}
    \boldsymbol{\phi} =(\boldsymbol{Z}^T\boldsymbol{W}\boldsymbol{Z})^{-1}\boldsymbol{Z}^T\boldsymbol{W}\boldsymbol{v} = \boldsymbol{R}\boldsymbol{v}.
\end{align}
Here $\boldsymbol{Z}$ is a $2^M \times (M+1)$ matrix with $1$s in the first column (to obtain $\phi_0$) and the binary representations\footnote{For example, the binary representation of $\s = \{1,3\}$ in an $M=4$-dimensional setting is $[1,0,1,0]$.} of the coalitions $\s \subseteq \M$ in the remaining columns. $\boldsymbol{W} = \operatorname{diag}(C, \w, C)$ is a $2^M \times 2^M$ diagonal matrix containing the Shapley kernel weights $k(M, |\s|)$ on the diagonal.
The $\w$ vector contains the $2^M-2$ finite Shapley kernel weights from \eqref{eq:ShapleyKernelWeights}. 
Finally, $\boldsymbol{v}$ is a column vector of size $2^M$ containing the contribution function values $v(\s)$. The $\s$ in $\boldsymbol{W}$ and $\boldsymbol{v}$ corresponds to the coalition of the corresponding row in $\boldsymbol{Z}$. The $\boldsymbol{R}$ matrix is independent of the explicands. When explaining $N_\text{explain}$ predictions (explicands), we can replace $\boldsymbol{v}$ with a $2^M \times N_\text{explain}$ matrix $\boldsymbol{V}$, where column $i$ contains the contribution functions for the $i$th explicand.

Computing either \eqref{eq:Shapley_value_formula} or \eqref{eq:ShapleyValuesDefWLS} is infeasible in higher dimensions as the number of coalitions $2^M$ grows exponentially with the number of features $M$. A common solution is to approximate the Shapley values by solving \eqref{eq:ShapleyValuesDefWLS} 
using a randomly sampled set $\D$ of coalitions $\s \subseteq \M$ (with replacement), instead of all the $2^M$ coalitions \citep{lundberg2017unified}.
The coalitions are sampled from a distribution proportional to the Shapley kernel weights in \eqref{eq:ShapleyKernelWeights}, while the empty and grand coalitions are always included and exempt from the sampling. 
The corresponding Shapley value approximation is 
\begin{align}
    \label{eq:ShapleyValuesDefWLSSolution_approx}
    \boldsymbol{\phi}_{\D} =(\boldsymbol{Z}_{\D}^T\boldsymbol{W}_{\D}\boldsymbol{Z}_{\D})^{-1}\boldsymbol{Z}_{\D}^T\boldsymbol{W}_{\D}\boldsymbol{v}_{\D} = \boldsymbol{R}_{\D}\boldsymbol{v}_{\D} 
\end{align}
where only the $N_{\text{coal}} = |\D|$ unique coalitions in $\D$ are used. If a coalition $\s$ is sampled $K$ times, then the corresponding weight in $\W_\D = \operatorname{diag}(C, \w_\D, C)$, is set to its sampling frequency, i.e., $w_\s = K$. In practice, the $w_\s$ values are often normalized for numerical stability. We refer to both \eqref{eq:ShapleyValuesDefWLSSolution} and the approximate solution in \eqref{eq:ShapleyValuesDefWLSSolution_approx} as KernelSHAP \citep{lundberg2017unified}.
\citet{williamson2020efficient} shows that the KernelSHAP approximation framework is consistent and asymptotically unbiased, while \citet{covert2021improving} demonstrates that it is empirically unbiased for even a modest number of coalitions.

Antithetic or \textit{paired} sampling is a simple variance reduction technique for Monte Carlo sampling \citep[Ch.~9]{kroese2013handbook}. In the KernelSHAP sampling setting, this has been utilized by always including $v(\thickbar{\mathcal{S}})$ whenever $v(\mathcal{S})$ is included in the computation, to significantly reduce the variance of the Shapley value estimates from \eqref{eq:ShapleyValuesDefWLSSolution_approx} \citep{covert2021improving, mitchell2022sampling}. Pairing coalitions stabilizes the sampling frequencies by ensuring that $\s$ and $\sbb$ consistently receive the same weight $w_\s$ in \eqref{eq:ShapleyValuesDefWLSSolution_approx}. 
However, due to randomness in the sampling procedure, 
the observed coalition weights may deviate from their nominal sampling probabilities -- for example, coalitions of the same size can easily end up with different weights.

To address this and further reduce the variance of the KernelSHAP estimator, \citet{Olsen2024improving} suggest to use reweighing strategies. 
The authors propose a simple correction to the frequency based WLS weights $w_{\s}$ used in \eqref{eq:ShapleyValuesDefWLSSolution_approx}. The corrected weights are calculated by normalizing the Shapley kernel weights in \eqref{eq:ShapleyKernelWeights} and conditioning on the coalition being sampled at least once, i.e.,
\begin{align}
    \tilde{p}_\s = \frac{p_\s}{\operatorname{Pr}(\text{$\s$ sampled at least once})} = \frac{p_\s}{1 - \operatorname{Pr}(\text{$\s$ not sampled})} = \frac{p_\s}{1 - (1 - p_\s)^L}, \label{eq:kernel_reweighter}
\end{align}
where $p_{\s} = k(M, |\s|)\big/\!\left(\sum_{q = 1}^{M-1} k(M, q)\binom{M}{q}\right)$ is the normalized version of the Shapley kernel weights and $L$ is the total number of sampled coalitions in $\D$ (including duplicates). 
 Thus, for a sampled coalition $\s$ the corrected weight $w_\s$ used in \eqref{eq:ShapleyValuesDefWLSSolution_approx} is 
\begin{align}
    \label{eq:weight_paired_c_kernel}
    w_\s \propto \frac{p_\s}{1 - (1 - p_\s)^{L}}.
\end{align}
Moreover, the authors describe a \textit{semi‑deterministic sampling} scheme implemented in the \pkg{shap} \proglang{Python} library \citep{shap}. 
The procedure essentially includes all coalitions of size $k$ when a fully random sample of size $N_{\text{coal}}$ would be expected to cover them, and samples the remaining coalitions randomly with probabilities proportional to their Shapley kernel weights in \eqref{eq:ShapleyKernelWeights}.

Extensive numerical experiments by \citet{Olsen2024improving} demonstrate that combining \textit{paired sampling} with the corrected Shapley kernel weights, both with and without the semi-deterministic sampling component, are superior as they 
match the accuracy of the standard KernelSHAP approximator using only $5\%$ to $50\%$ of the coalitions.


\subsection{Estimation paradigms} \label{sec:background:paradigms}
The model-agnostic approaches used to estimate the conditional contribution function in \eqref{eq:ContributionFunc} can be categorized into two main paradigms: The \textit{Monte Carlo} and \textit{regression} paradigms. 

In the Monte Carlo paradigm, Monte Carlo integration is used to estimate $v(\mathcal{S})$:
\begin{align}
    \label{eq:KerSHAPConditionalFunction}
    v(\mathcal{S})   
    =
    \E\left[ f(\boldsymbol{x}_{\thickbar{\mathcal{S}}}, \boldsymbol{x}_{\mathcal{S}}) | \boldsymbol{x}_{\mathcal{S}} = \boldsymbol{x}_{\mathcal{S}}^* \right] 
    \approx
    \frac{1}{K} \sum_{k=1}^K f(\boldsymbol{x}_{\thickbar{\mathcal{S}}}^{(k)}, \boldsymbol{x}_{\mathcal{S}}^*) 
    =    
    \hat{v}(\mathcal{S}),
\end{align}
where $\boldsymbol{x}_{\thickbar{\mathcal{S}}}^{(k)} \sim p(\boldsymbol{x}_{\thickbar{\mathcal{S}}} | \boldsymbol{x}_{\mathcal{S}} = \boldsymbol{x}_{\mathcal{S}}^*)$, for $k=1,2,\dots,K$, and $K$ is the number of Monte Carlo samples. 
The estimated $v(\mathcal{S})$ can then be inserted into \eqref{eq:Shapley_value_formula}, \eqref{eq:ShapleyValuesDefWLSSolution}, or \eqref{eq:ShapleyValuesDefWLSSolution_approx} to approximate the Shapley values. 
To obtain accurate conditional Shapley values, we need to generate Monte Carlo samples that follow the conditional distribution of the data. 
This distribution is generally not known and needs to be estimated based on the training data. Approaches for estimating those are discussed in Section \ref{sec:background:approaches}.

The regression paradigm uses the fact that the conditional expectation is the minimizer of the mean squared error loss function:
\begin{align}
\label{eq:ContributionFunc:Regression}
\begin{split}
    v(\s) &= \E\left[ f(\xsb, \xs) | \xs = \xss \right] = \argmin_c \E\left[(f(\xsb, \xs) - c)^2 | \xs = \xss\right].
\end{split}
\end{align}
Thus, any regression model $g_\s(\xs)$, which is fitted with the mean squared error loss function as the objective function will approximate \eqref{eq:ContributionFunc:Regression}, obtaining an alternative estimator $\hat{v}(\s)$ of $v(\s)$ \citep{frye_shapley-based_2020, williamson2020efficient, olsen2024comparative}. The accuracy of the approximation will depend on the form of the predictive model $f(\x)$, the flexibility of the regression model $g_\s(\xs)$, and the optimization routine. We can either train a separate regression model $g_\s(\xs)$ for each $\s$ or we can train a single regression model $g(\tilde{\x}_\s)$ which approximates the contribution function $v(\s)$ for all $\s$ simultaneously. 
Here $\tilde{\x}_\s$ is an augmented version of $\xs$ with fixed-length representation (see \textit{Regression Surrogate} in Section~\ref{sec:background:approaches}).

\begin{figure}[t!]
\centering
\includegraphics[width=0.7\linewidth]{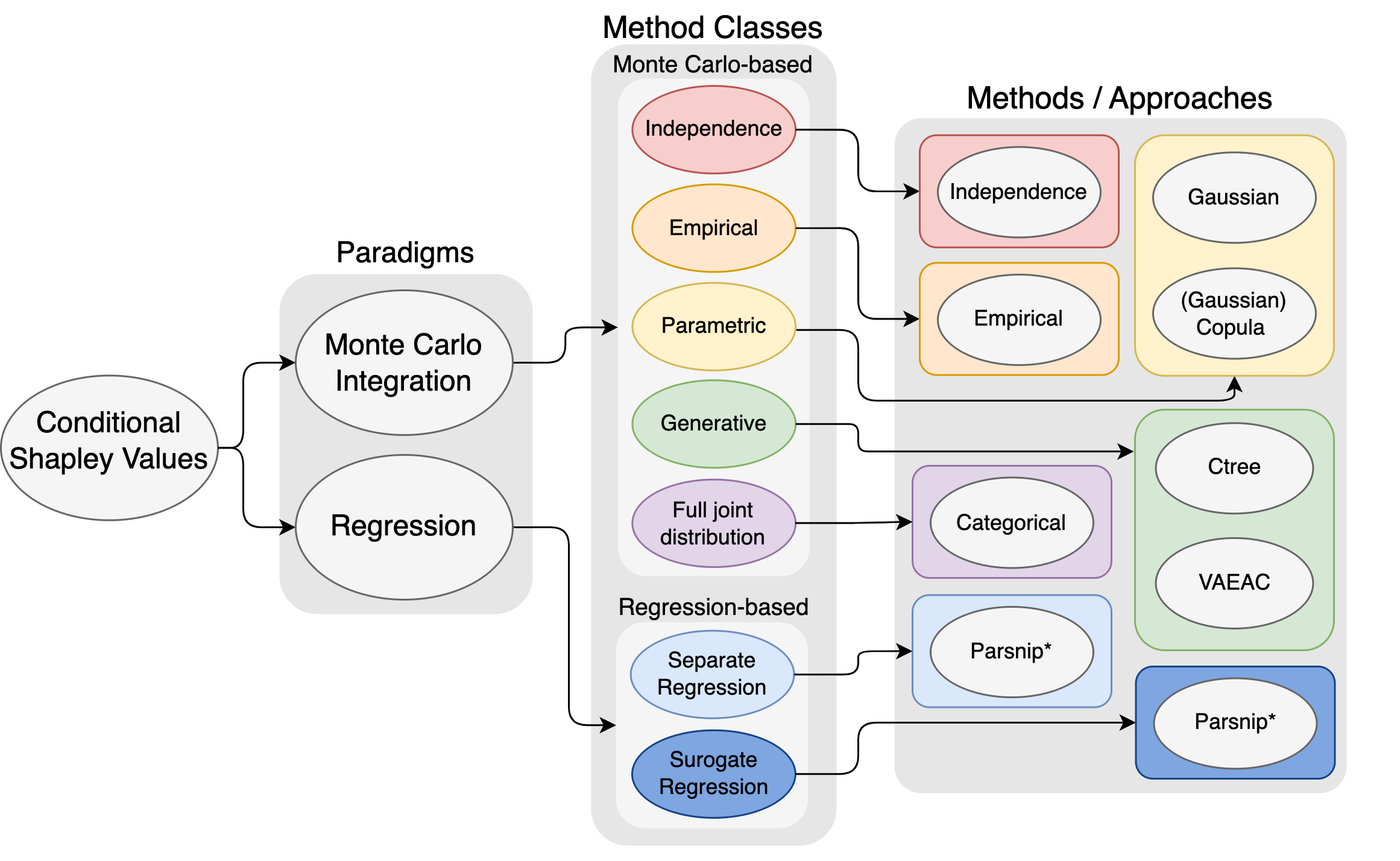}
\caption{Schematic overview of the available approaches in \pkg{shapr} for computing conditional Shapley value explanations. \textit{Parsnip*} indicates that the regression-based approaches can use any regression methods available in the \pkg{parsnip} package \citep{Parsnip}, including user-specified methods. The \pkg{shapr} package also implements the \code{timeseries} approach, which is specifically designed for time series data and not applicable to regular tabular datasets. See \citet[Sec.~3.4]{jullum2021groupshapley} for details.}
\label{fig:Shaple_values_map} 
\end{figure}

\subsection{Estimation approaches} \label{sec:background:approaches}
Below we give brief introductions to a wide range of approaches for estimating $v(\s)$ with both paradigms described in Section \ref{sec:background:paradigms}, all of which are implemented in the \pkg{shapr} package. 
Unless ``Regression'' is explicitly mentioned in the approach name, the approaches described below follow the Monte Carlo paradigm.
Figure \ref{fig:Shaple_values_map} provides a schematic overview of the approaches.

In addition to the model-agnostic approaches described below, \pkg{shapr} also includes the \linebreak\code{timeseries} approach. This is specifically designed for time series data and not suitable for standard tabular datasets. For more details of that special case, we refer to \citet[Sec.~3.4]{jullum2021groupshapley}.

\subsubsection{Independence} \label{sec:background:approaches:independence}
\citet{lundberg2017unified} avoided estimating the complex conditional distributions by implicitly assuming feature independence. 
While this often leads to incorrect conditional Shapley value explanations for real-world data \citep{aas2019explaining,frye_shapley-based_2020} (and is therefore not recommended) the method is included for reference.
In the \independence\ approach, the conditional distribution $p(\xsb|\xs)$ simplifies to $p(\xsb)$, and the corresponding Shapley values are the marginal Shapley values obtained by using the $v_{\text{marg}}(\mathcal{S})$ in \eqref{eq:ContributionFuncMarg}. 
The Monte Carlo samples $\boldsymbol{x}_{\thickbar{\mathcal{S}}}^{(k)} \sim p(\boldsymbol{x}_{\thickbar{\mathcal{S}}})$ in \eqref{eq:KerSHAPConditionalFunction} are generated by randomly sampling observations from the training data; thus, no modeling is needed.
For this reason, this is usually a very fast approach.

\subsubsection{Empirical} \label{sec:background:approaches:empirical}
A natural way to account for feature dependence when sampling from the training data is to sample observations similar to $\boldsymbol{x}^*_{\s}$, rather than sampling entirely at random as done in the \independence\ approach. 
This is the idea behind the \empirical\ method described in \citet{aas2019explaining}.



A scaled version of the Mahalanobis distance $D_\s$ is used to calculate a distance $D_\s(\x^*, \x^{[i]})$ between the explicand $\x^*$ and every training instance $\x^{[i]}$, for the features in $\s$. Then, a Gaussian distribution kernel is used to convert the distance into a weight $w_\s(\x^*, \x^{[i]})$. All the weights are sorted in increasing order with $\x^{\set{k}}$ having the $k$th largest value. Finally, \eqref{eq:ContributionFunc} is approximated by a weighted version of \eqref{eq:KerSHAPConditionalFunction}:
\begin{align}
    \hat{v}(\mathcal{S}) 
    =
    \sum_{k=1}^{K^*}
    \frac{
    w_\s(\x^*, \x^{\set{k}})
    f(\boldsymbol{x}_{\thickbar{\mathcal{S}}}^{\set{k}}, \boldsymbol{x}_{\mathcal{S}}^*)
    }{
    \sum_{k=1}^{K^*}
    w_\s(\x^*, \x^{\set{k}})}. \label{eq:weightedmean}
\end{align}
The $K^*$ here is typically set such that, for instance, $\eta=95\%$ of the total weight across the training set is accounted for: $K^* = \min_{L \in \N} \big\{\sum_{k=1}^L w_\s(\x^*, \x^{\set{k}}) \big/ \sum_{i=1}^{N_\text{train}} w_\s(\x^*, \x^{[i]}) > \eta \big\}$.

\subsubsection{Gaussian} \label{sec:background:approaches:gaussian}
For well-behaved unimodal data, a natural approach is to assume that $\boldsymbol{x}$ stems from a multivariate Gaussian distribution with mean $\bmu$ and covariance matrix $\bSigma$.
In the present context, this method was first suggested by \citet{aas2019explaining}.

A classic and beneficial property of this assumption is that the conditional distributions are also multivariate Gaussian. 
That is, if $p(\x) = p(\xs, \xsb) = \mathcal{N}_M(\bmu, \bSigma)$, where $\bmu = [\bmu_{\s}, \bmu_{\sbb}]^T$ and $\boldsymbol{\Sigma} = \Big[\begin{smallmatrix} \bSigma_{\s\s} & \bSigma_{\s\sbb} \\ \bSigma_{\sbb\s} & \bSigma_{\sbb\sbb} \end{smallmatrix}\Big]$, then, $p(\xsb| \xs = \xss) = \mathcal{N}_{|\sbb|} (\bmu_{\sbb|\s}, \bSigma_{\sbb|\s})$, where $\bmu_{\sbb|\s} = \bmu_{\sbb} + \bSigma_{\sbb\s} \bSigma_{\s\s}^{-1}(\xss - \bmu_{\s})$ and $\bSigma_{\sbb|\s} = \bSigma_{\sbb\sbb} - \bSigma_{\sbb\s}\bSigma_{\s\s}^{-1}\bSigma_{\s\sbb}$. The parameters $\bmu$ and $\bSigma$ are easily estimated using, respectively, the sample mean and covariance matrix of the training data. 
Finally, \eqref{eq:KerSHAPConditionalFunction} is used to estimate $v(\mathcal{S})$ with samples $\boldsymbol{x}_{\thickbar{\mathcal{S}}}^{(k)}$ from $p(\boldsymbol{x}_{\thickbar{\mathcal{S}}} | \boldsymbol{x}_{\mathcal{S}} = \boldsymbol{x}_{\mathcal{S}}^*)$, for $k=1,2,\dots,K$.

\subsubsection{Gaussian copula} \label{sec:background:approaches:copula}
A generalization of the \gaussian\ approach, also proposed by \citet{aas2019explaining}, is to use a Gaussian copula. That is, to represent the marginals of the features by their empirical distributions and then model the dependence structure by a Gaussian distribution. 
This \code{copula} approach generates the $K$ conditional Monte Carlo samples $\boldsymbol{x}_{\thickbar{\mathcal{S}}}^{(k)} \sim p(\boldsymbol{x}_{\thickbar{\mathcal{S}}} | \boldsymbol{x}_{\mathcal{S}} = \boldsymbol{x}_{\mathcal{S}}^*)$ with the following routine:
\begin{enumerate}
    \item Convert each marginal $x_j$ of the feature distribution $\boldsymbol{x}$ to a Gaussian distributed variable $v_j$ by $v_j = \Phi^{-1}(\hat{F}(x_j))$, where $\hat{F}(x_j)$ is the empirical distribution function of $x_j$.
    \item Assume that $\boldsymbol{v}$ is distributed according to a multivariate Gaussian, and sample from the conditional distribution $p(\boldsymbol{v}_{\sbb} | \boldsymbol{v}_{\s} = \boldsymbol{v}_{\s}^*)$ using the method described for the \gaussian\ approach.
    \item Convert the marginals $v_j$ in the conditional distribution to the original distribution using $\hat{x}_j = \hat{F}_j^{-1}(\Phi(v_j))$.
\end{enumerate}

\subsubsection{Ctree} \label{sec:background:approaches:ctree}
\citet{redelmeier2020} compute conditional Shapley values by modeling the dependence structure between the features with conditional inference trees (\textit{ctree}) to then sample from these, i.e., using a generative method. 
A \textit{ctree} is a type of recursive partitioning algorithm that builds trees recursively by doing binary splits on features until a stopping criterion is satisfied \citep{HothornCtree}. 
The process is sequential, where the splitting feature is chosen first using statistical significance tests, and then the splitting point is chosen using a splitting criterion. 
The \textit{ctree} algorithm is independent of the dimension of the response, which in our case is $\xsb$. 
The input features are $\xs$, which varies in dimension based on the coalition $\s$. 
That is, for each coalition $\s$, a \textit{ctree} with $\xs$ as the features and $\xsb$ as the response is fitted to the training data. 
For a given $\xss$, the \ctree\ approach finds the corresponding leaf node and samples $K$ observations with replacement from the $\xsb$ part of the training observations in the same node to generate the conditional Monte Carlo samples $\boldsymbol{x}_{\thickbar{\mathcal{S}}}^{(k)} \sim p(\boldsymbol{x}_{\thickbar{\mathcal{S}}} | \boldsymbol{x}_{\mathcal{S}} = \boldsymbol{x}_{\mathcal{S}}^*)$. 
Whenever $K$ is larger than the number of samples in the leaf node, the \ctree\ approach uses all unique samples in the leaf node. It computes the $v(\s)$ as a weighted mean similar to \eqref{eq:weightedmean} instead of by \eqref{eq:KerSHAPConditionalFunction}.

\subsubsection{VAEAC} \label{sec:background:approaches:vaeac}
The \vaeac\ approach is another generative method for estimating $v(\mathcal{S})$. 
\citet{Olsen2022} use a type of variational autoencoder called \textit{vaeac} \citep{ivanov_variational_2018} to generate the conditional Monte Carlo samples. Briefly stated, the original variational autoencoder \citep{kingma2014autoencoding,Kingma2019AnIT,pmlr-v32-rezende14} gives a probabilistic representation of the true unknown distribution $p(\boldsymbol{x})$. The \textit{vaeac} model extends this methodology to all conditional distributions $p(\xsb | \xs = \xss)$ simultaneously. That is, a single \textit{vaeac} model can generate Monte Carlo samples $\x_{\thickbar{\mathcal{S}}}^{(k)} \sim p(\boldsymbol{x}_{\thickbar{\mathcal{S}}} | \boldsymbol{x}_{\mathcal{S}} = \boldsymbol{x}_{\mathcal{S}}^*)$ for all coalitions $\mathcal{S} \subseteq \M$. It is advantageous to only have to fit a single model for all coalitions, as the number of coalitions increases exponentially with the number of features. 
In contrast, \ctree\ trains $2^M-2$ different models, eventually becoming computationally intractable for large $M$. 
The \vaeac\ approach trains a \textit{vaeac} model by maximizing a variational lower bound, which conceptually corresponds to artificially masking features, and then trying to reproduce them using a probabilistic representation. 
In deployment, it considers the unconditional features $\xsb$ as masked features to be imputed.

\subsubsection{Categorical}
\label{sec:categorical}
When the features are all categorical, we can estimate the conditional expectations using basic statistical theory. 
That is, all marginal and joint probabilities are estimated by simply counting
the frequencies of each feature, feature pair, feature triplets, etc. 
Then all conditional probabilities can be computed by dividing the full joint distributions by the necessary lower-order joint distributions.
Since this provides a complete tabular description of all conditional distributions $p(\boldsymbol{x}_{\thickbar{\mathcal{S}}}|\boldsymbol{x}_{\mathcal{S}})$, we can easily \textit{compute} the 
$v(\mathcal{S})$ through 
\begin{align}
v(\mathcal{S}) = \E[f(\boldsymbol{x})|\boldsymbol{x}_{\mathcal{S}}=\boldsymbol{x}_{\mathcal{S}}^*] = \sum_{\z \in \mathcal{Z}} f(\boldsymbol{x}_{\mathcal{S}}^*,\z) p(\boldsymbol{x}_{\thickbar{\mathcal{S}}}=z|\boldsymbol{x}_{\mathcal{S}} = \boldsymbol{x}_{\mathcal{S}}^*),
\end{align}
where $\mathcal{Z}$ is the finite sample space of $\boldsymbol{x}_{\thickbar{\mathcal{S}}}|\boldsymbol{x}_{\mathcal{S}}$.
Even though this is an exact computation of the expectation, given the estimated marginal and joint distributions, we group the method together
with the Monte Carlo approaches since the computation of the expectations takes the same form.
For computational reasons, the approach is most relevant when the cardinality (number of factor levels) of the features is not too large. 

\subsubsection{Regression separate} \label{sec:background:approaches:regression_separate}
Turning to the regression paradigm, the \code{regression\_separate} approach estimates $v(\mathcal{S}) = \E\left[ f(\x) | \xs = \xss \right]$ separately for each coalition $\s$ using regression \citep{olsen2024comparative}. 
Let $\x^{[i]}$ denote the $i$th $M$-dimensional input and $y^{[i]}$ the associated response of the training data. For each $\s$, define the coalition dataset
\begin{align*}
            \mathcal{X}_\mathcal{S} 
            =
            \{\x_\s^{[i]}, f(\underbrace{\x_\sbb^{[i]}, \x_\s^{[i]}}_{\x^{[i]}})\}_{i=1}^{N_{\text{train}}}
            =
            \{\x_\s^{[i]}, \underbrace{f(\x^{[i]})}_{z^{[i]}}\}_{i=1}^{N_{\text{train}}}
            =
            \{\x_\s^{[i]}, z^{[i]}\}_{i=1}^{N_{\text{train}}}.
\end{align*}

For each $\mathcal{X}_\s$, we train a regression model $g_\s(\xs)$ with the mean squared error loss function, which has the property of aiming at estimating precisely $\E[z|\xs] = \E[f(\xsb, \xs)|\xs]$. To get estimates of $v(\mathcal{S})$ for a specific explicand $\boldsymbol{x}^*$, the estimated regression model $g_\s(\xs)$ is simply evaluated at $\boldsymbol{x}_{\s}^*$.
See, e.g., \citet[Sec.~3.5]{olsen2024comparative} for more details.

\subsubsection{Regression surrogate} \label{sec:background:approaches:regression_surrogate}
A drawback of the \code{separate\_regression} approach is that training a separate regression model $g_\s(\x_\s)$ for each coalition $\s$ becomes infeasible as the number of features $M$ increases. Additionally, the separate training of each $g_\s$ from scratch prevents them from borrowing strength from each other, even though the regression problems may be similar for similar coalitions. The \code{surrogate\_regression} approach addresses these limitations by training a single regression model $g$ on an augmented dataset that handles all coalitions simultaneously.

The surrogate model $g$ must account for the varying size of $\xs$ for each coalition $\s$. \citet[Sec.~3.6]{olsen2024comparative} solve this problem by creating a fixed-length representation $\tilde{\x}_\s$ of $\xs$ for all coalitions $\s$. The $\tilde{\x}_\s$ representation is of length $2M$, with the two halves being the fixed-length representations of $\xs$ and $\s$, respectively. The latter is included to enable the regression model $g$ to distinguish between different coalitions. The training dataset is augmented by systematically applying these fixed-length representations to each training observation. The surrogate regression model $g$ is then fitted to this augmented dataset using the mean squared loss function, as done for the \code{separate\_regression} approach. This enables $g$ to estimate the contribution function $v(\s)$ for all coalitions $\s \subseteq \mathcal{M}$ simultaneously.
See, e.g., \citet[Sec.~3.6]{olsen2024comparative} for more details.

\subsection{Iterative KernelSHAP and convergence detection} 
\label{sec:iterative}

Since it is computationally intensive to estimate many $v(\mathcal{S})$ with most of the methods described in Section \ref{sec:background:approaches}, we aim to use only as many as needed to achieve sufficiently accurate Shapley value estimates. 
A simple solution to this is to iteratively add more $v(\mathcal{S})$ values, and stop once the desired accuracy level is reached.
This can be done by initially sampling a set of coalitions and then using bootstrapping \citep[Sec.~4.2]{goldwasser2024stabilizing} to estimate the variance of the approximated Shapley values. 
A convergence criterion is used to determine if the variances of the Shapley values are sufficiently small. 
If the variances are too large, we may estimate the number of required samples to reach convergence and then increase the number of coalitions accordingly. The process is repeated until the variances are below the convergence threshold. 

\citet{covert2021improving} outline such an algorithm with a convergence criterion for computing the Shapley values of a single explicand $\boldsymbol{x}^*$. 
We suggest modifying their criterion to work for multiple observations with a simple median across the explicands, leaving us with
\begin{align}
    \underset{i}{\text{median}}\left(\frac{\max_j \text{sd}(\phi_{ij})}{\max_j \phi_{ij} - \min_j \phi_{ij}}\right) < t,
    \label{eq:convcrit}
\end{align} 
where $\phi_{ij}$ denotes the estimated Shapley value of feature $j$ for explicand $i$, and $\text{sd}(\phi_{ij})$ is its (bootstrap) estimated standard deviation. 
The convergence threshold value $t$ may be set to a small number (\citet{covert2021improving} suggest 0.01).

\subsection[MSEv evaluation criterion]{$\operatorname{MSE}_{v}$ evaluation criterion}
\label{sec:MSEv}

To evaluate and rank Shapley values from different approaches in Section \ref{sec:background:approaches}, \citet{frye_shapley-based_2020} proposed to use the mean squared error across $N_\text{explain}$ observations $\x^{[1]},\ldots,\x^{[N_{\text{explain}}]}$:
\begin{align}
    \label{eq:MSE_v}
    \operatorname{MSE}_{v}(\text{method } \texttt{q}) 
    =
     \frac{1}{N_{\text{coal}}} \sum_{\s} \frac{1}{N_\text{explain}} \sum_{i=1}^{N_\text{explain}} \left( f(\boldsymbol{x}^{[i]}) - \hat{v}_{\texttt{q}}(\s, \boldsymbol{x}^{[i]})\right)^2\!,
\end{align}
where $N_{\text{coal}}$ is the number of used coalitions and $\hat{v}_{\texttt{q}}(\s, \boldsymbol{x}^{[i]})$ denotes the estimated contribution function using method $\texttt{q}$, evaluated at $\boldsymbol{x}^{[i]}$. 

The motivation behind the $\operatorname{MSE}_{v}$ criterion is that $\mathbb{E}_\s\mathbb{E}_{\x} (v_{\texttt{true}}(\s, \boldsymbol{x}) - \hat{v}_{\texttt{q}}(\s, \boldsymbol{x}))^2$ can be decomposed as
\begin{align}
    \begin{split}
    \mathbb{E}_\s\mathbb{E}_{\x} (v_{\texttt{true}}(\s, \boldsymbol{x})- \hat{v}_{\texttt{q}}(\s, \boldsymbol{x}))^2 
    &=
    \mathbb{E}_\s\mathbb{E}_{\x} (f(\x) - \hat{v}_{\texttt{q}}(\s, \boldsymbol{x}))^2 - \mathbb{E}_\s\mathbb{E}_{\x} (f(\x)-v_{\texttt{true}}(\s, \boldsymbol{x}))^2,
    \end{split} \notag
\end{align}
see \citet[Appendix A]{covert2020understanding}. The first term on the right-hand side can be estimated by \eqref{eq:MSE_v}, while the second term is a fixed (unknown) constant not influenced by the approach \texttt{q}. Thus, a low value of $\operatorname{MSE}_{v}$ in \eqref{eq:MSE_v} indicates that the estimated contribution function $\hat{v}_{\texttt{q}}$ is closer to the true counterpart $v_{\texttt{true}}$ than a high value.

The $\operatorname{MSE}_{v}$ evaluation criterion is helpful as it does not need knowledge about the true explanations. Thus, we can apply it to real-world datasets where the true Shapley values are unknown. 
However, the criterion has two drawbacks. First, we can only use it to rank the methods and not assess their closeness to the optimum since the minimum value of the $\operatorname{MSE}_{v}$ criterion is unknown. Second, the criterion evaluates the contribution functions rather than the Shapley values. Thus, the estimates for $v(\s)$ can undershoot and overshoot for different coalitions, but these errors might cancel each other out in the Shapley value formula \eqref{eq:Shapley_value_formula}.

Nevertheless, in a comprehensive simulation study, \citet[Figure~11]{olsen2024comparative} observe a relatively linear relationship between the $\operatorname{MSE}_{v}$ evaluation scores and the mean absolute error ($\operatorname{MAE}$) between the estimated and true Shapley values. That is, a method that achieves a low $\operatorname{MSE}_{v}$ score also tends to obtain a low $\operatorname{MAE}$ score, and vice versa. We therefore consider \eqref{eq:MSE_v} to be a decent evaluation metric for Shapley value estimates.

\section[The shapr R-package]{The \pkg{shapr} R-package}
\label{shapr:R}

The philosophy behind the \pkg{shapr} package is to provide a minimal set of user functions, which have good default settings for most use cases and are easy to use.
Furthermore, we aim to offer extensive flexibility for advanced users to define both \textit{how} computations should be performed, the desired level of accuracy, and the type of Shapley values that should be computed.

The \pkg{shapr} package is available from the Comprehensive R Archive Network (CRAN) at
\href{https://CRAN.R-project.org/package=shapr}{CRAN.R-project.org/package=shapr} or from the package's 
GitHub repository at \linebreak \href{https://github.com/NorskRegnesentral/shapr}{github.com/NorskRegnesentral/shapr}.
Documentation and the full suite of four vignettes showcasing the extensive functionality of the package are also easily accessible through \pkg{shapr}'s \pkg{pkgdown} \citep{pkgdown} online documentation at 
\href{https://norskregnesentral.github.io/shapr/}{norskregnesentral.github.io/shapr/}.
\pkg{shapr} can be installed directly from CRAN by:
\begin{CodeChunk}
\begin{CodeInput}
R> install.packages("shapr")
\end{CodeInput}
\end{CodeChunk}

Below we describe the basic usage, functionality, and flexibility of the \pkg{shapr} \proglang{R} package for explaining predictions with conditional Shapley values, and showcase its usage with a few practical code examples.

\subsection{Basic usage} 
\label{shapr:R:basic_usage}
The \code{explain()} function is the main function of the \pkg{shapr} \proglang{R} package. 
The function is used to set up and execute all computations for producing Shapley value-based explanations for a set of predictions from a model, and takes the following form:

\begin{CodeChunk}
\begin{CodeInput}
explain <- function(model,
                    x_explain,
                    x_train,
                    approach,
                    phi0,
                    iterative = NULL,
                    max_n_coalitions = NULL,
                    group = NULL,
                    n_MC_samples = 1e3,
                    seed = NULL,
                    verbose = "basic",
                    predict_model = NULL,
                    get_model_specs = NULL,
                    prev_shapr_object = NULL,
                    asymmetric = FALSE,
                    causal_ordering = NULL,
                    confounding = NULL,
                    extra_computation_args = list(),
                    iterative_args = list(),
                    output_args = list(),
                    ...)
\end{CodeInput}
\end{CodeChunk}

The many input arguments reflect the package’s flexibility. 
Most of the arguments are related to \textit{how} the estimation and computation of the Shapley values should be carried out and have good default values applicable for standard use cases.
Only the following five arguments need to be set by the user:
\begin{itemize}
    \item \code{model}: The model object whose predictions we want to explain.
    \item \code{x_explain}: The dataset whose predictions we want to explain.
    \item \code{x_train}: The dataset used to estimate the $v(\mathcal{S})$.
    \item \code{approach}: The approach used to estimate the $v(\mathcal{S})$. Allowed values are: \code{"gaussian"}, \code{"copula"}, \code{"empirical"}, \code{"ctree"}, \code{"vaeac"}, \code{"categorical"}, \code{"timeseries"},\newline \code{"independence"}, \code{"regression_separate"}, and \code{"regression_surrogate"}, corresponding to the approaches described in Section \ref{sec:background:approaches}.
    \item \code{phi0}: The value to use for $\phi_0=\mathbb{E}[f(\boldsymbol{x})]$. For (approximately) unbiased models, $f(\cdot)$, we recommend setting this to the mean of the response used to fit the model.
\end{itemize}

The output of \code{explain()} is an object of class \code{(shapr, list)} and contains the following elements:
\begin{itemize}
    \item \code{shapley_values_est}: Table with the estimated Shapley values.
    \item \code{shapley_values_sd}: Table with standard deviations of the estimated Shapley values (with respect to the sampling uncertainty in KernelSHAP described in Section \ref{sec:background:kernelshap}).
    \item \code{pred_explain}: The prediction outcome for the observations we are explaining.
    \item \code{MSEv}: The $\operatorname{MSE}_{v}$ evaluation criterion for the approach as described in Section \ref{sec:MSEv}.
    \item \code{iterative_results}: Details on the iterative estimation, its convergence, and the intermediate Shapley value estimates.
    \item \code{saving_path}: The path where the iterative results are saved. 
    \item \code{internal}: Various additional information about the performed computations.
    \item \code{timing}: The time spent executing the various parts of the function call.
\end{itemize}

The \code{shapr} class provides three methods (attribution functions):

\begin{itemize}
    \item \code{print.shapr()}: Displays a table of the computed Shapley values by default. The optional \code{what} argument can be used to show other components, such as standard deviations, MSE estimates, and timing summaries.
    \item \code{plot.shapr()}: Creates various visualizations of the estimated Shapley values using \pkg{ggplot2} \citep{ggplot2}. The function returns a \code{(gg, ggplot)} object, enabling full use of \pkg{ggplot2}'s customization features. By default, bar plots are produced for each explicand. Alternative visualizations, such as \code{"waterfall"}, \code{"scatter"}, and \code{"beeswarm"} plots, can be specified via the \code{plot_type} argument.
    \item \code{summary.shapr()}: Returns a \code{summary.shapr} object which provides convenient access to information related to the Shapley value computation. 
    It includes a dedicated \code{print} method that displays a cleanly formatted summary to the console.
\end{itemize}
In addition, \code{get_results()}, which is called internally by \code{summary.shapr()} to extract the results from \code{shapr} objects, can be called directly to retrieve specific objects via its \code{what} argument.

\subsection{Package functionality} \label{sec:package_functinality}

The \pkg{shapr} package provides a wide range of functionality and flexibility related to the \textit{type} of Shapley values computed, the way they are computed (computationally), and the feedback given to the user during the potentially long-running computations.
Below, we describe the most important functionality and how it can be controlled and specified by the user of the package.

\subsubsection{Supported models}

\pkg{shapr} provides native support for explaining predictions from models fit using \code{stats::lm}, \code{stats::glm}, \code{ranger::ranger} \citep{ranger}, \code{mgcv::gam} \citep{mgcv}, and \code{xgboost::xgboost}/\code{xgboost::xgb.train}  \citep{xgboost}, in addition to the hundreds of models available in the popular \pkg{parsnip} package \citep{Parsnip} when fit through the associated \pkg{workflows}  \citep{workflows} package. 
The latter is recommended when using models that require one-hot encoding of categorical features, such as, e.g., \pkg{xgboost} \citep{xgboost}, as \pkg{workflows} handles this transformation automatically, making the process easier and less error-prone for the user.
We refer the reader to the \href{https://workflows.tidymodels.org/index.html}{\pkg{pkgdown} website} of \pkg{workflows} for details on how to fit models with the \pkg{workflows} package.
For all of these models, the user needs only to pass the model object to \code{explain()}. The \pkg{shapr} package will then check consistency between the \code{model} object and the data in \code{x_train} and \code{x_explain}, and generate the necessary predictions from the model when needed.
Moreover, any other predictive model that outputs a single numeric value can be explained using \code{explain()} by supplying a custom prediction function through the \code{predict_model} argument. The \code{predict_model} function should take \code{model} and a \code{data.frame/data.table} as arguments and return the associated predictions as a numeric vector.  
To enable feature and model checking, an additional function can be supplied via the \code{get_model_specs} argument.
The required structure of this function is provided in Appendix \ref{app:get_model_specs}.

\subsubsection{Group-wise Shapley values}

In certain cases, particularly when the number of features is large, it may be more effective and informative to explain predictions using groups of features rather than individual ones. 
By defining coalitions of feature \textit{groups} rather than single features, the computational complexity of the Shapley value computations may also be significantly reduced. 
Group-wise Shapley values can be computed with \code{explain()} by supplying a (named) list of feature vector names to \code{group}. 
For further intuition and real-world examples of group-wise Shapley values, see \citet{jullum2021groupshapley, au2022grouped}.

\subsubsection{Combining approaches}

In some use cases, it may be beneficial to use different approaches for estimating $v(\mathcal{S})$ for different coalitions $\mathcal{S}$. 
As suggested by \citet[Sec.~3.4]{aas2019explaining}, \citet{izbicki2017converting} among others, simpler models may be more appropriate for estimating conditional distributions on the form $p(\boldsymbol{x}_\text{few}|\boldsymbol{x}_\text{many})$ than 
$p(\boldsymbol{x}_\text{many}|\boldsymbol{x}_\text{few})$, where $\boldsymbol{x}_\text{many}$ is of (much) higher dimension than $\boldsymbol{x}_\text{few}$.
For instance, a Gaussian distribution may sometimes be a fast and reasonable approach to estimate $p(\boldsymbol{x}_\text{few}|\boldsymbol{x}_\text{many})$ (i.e., when $|\mathcal{S}|$ is large), while being a poor approximation approach to estimate $p(\boldsymbol{x}_\text{many}|\boldsymbol{x}_\text{few})$ (i.e., when $|\mathcal{S}|$ is small). In the latter case, the more flexible \textit{ctree} or \textit{vaeac} approaches may be more appropriate.
Combining approaches can be done in \code{explain()} by setting \code{approach} equal to a character vector of length $M-1$, where $M$ is the number of features. In that case, the $j$-th element specifies the approach to use for $v(\mathcal{S})$ when $|\mathcal{S}| = j$.

\subsubsection{Causal and asymmetric Shapley values}

Asymmetric \citep{frye2020asymmetric} and Causal \citep{heskes2020causal} Shapley values are modified Shapley value frameworks that aim at incorporating causal knowledge into the explanations. 
The frameworks require knowledge or assumptions about a (partial) causal ordering of the features or feature-groups and whether confounders exist for the features/groups in the model.
All of this must be specified through the arguments \code{asymmetric}, \code{causal_ordering}, and \code{confounding}. They are all disabled by default. 
Section \ref{sec:asymmetric_and_causal_shapley_values} describes the methodologies and how to compute such Shapley values with the \code{explain()} function.

\subsubsection{Flexible estimation with the regression paradigm}

As described in Sections \ref{sec:background:paradigms}--\ref{sec:background:approaches}, the \code{separate\_regression} approach trains a new regression model to estimate the $v(\mathcal{S})$ values for each coalition $\mathcal{S}$, while the \code{surrogate\_regression} approach trains a single regression model for all coalitions $\s$.
For these approaches, \pkg{shapr} allows using any regression model in the popular \pkg{parsnip} package. 
These regression models can also preprocess the data and be tuned to optimize their performance by leveraging other packages in the \pkg{tidymodels} framework \citep{tidymodels}, such as \pkg{recipes}, \pkg{tune}, and \pkg{workflows} through the ellipsis arguments \code{regression.model}, \code{regression.tune\_values}, \code{regression.vfold\_cv\_para}, \code{regression.recipe\_func} and \code{regression.surrogate\_n\_comb} of \code{explain()}.
The default regression model for both regression-based approaches (\code{approach} \code{=} \code{"regression_separate"} and \code{approach} \code{=} \code{"regression_surrogate"}) is a standard linear regression model. 
For more details, see the regression vignette
``Shapley value explanations using the regression paradigm'' available through \code{vignette("regression","shapr")} or the \href{https://norskregnesentral.github.io/shapr/articles/regression.html}{online documentation}.
We also provide an example of the \code{regression_surrogate} approach in Section \ref{sec:conditional_shapley_values:examples}.


\subsubsection{Shapley values for forecasting models}

The \code{explain()} function explains single-outcome predictions.
However, in some cases, particularly in time series forecasting settings, we are interested in explaining multiple outcomes from the same observations.
While in principle, this can be achieved by calling \code{explain()} multiple times with custom \code{predict_model}s as described above, it is not computationally efficient because the computationally intensive part (modeling of the conditional distributions) is repeated unnecessarily.
Therefore, \pkg{shapr} offers a dedicated function, \code{explain_forecast()}, designed for the specific use-case of explaining forecasts from time series models at one or more future time steps. The usage and functionality of this function is described in Section~\ref{sec:conditional_shapley_values_forecast}.

\subsubsection{Improved KernelSHAP}

In Section~\ref{sec:background:kernelshap}, we discussed how paired sampling, variance-reducing reweighting, and semi-deterministic sampling 
 of coalitions significantly improve the efficiency of KernelSHAP. 
 The \pkg{shapr} package uses paired sampling (unless \code{asymmetric = TRUE}) and the reweighting method in \eqref{eq:kernel_reweighter} by default. 
 Although not recommended, these can be disabled by setting \newline \code{paired_shap_sampling = FALSE} and \code{kernelSHAP_reweighting = "none"} in the list passed to the \code{extra_computation_args} argument of \code{explain()}. 
 Semi-deterministic sampling is disabled by default but can be enabled with \code{semi_deterministic_sampling = TRUE} in the same list. 
 A few other reweighting strategies are also available. 
For a complete overview of all available options, as well as other computational defaults and settings that can be passed via \code{extra_computation_args}, see the help file of \code{get_extra_comp_args_default}.

\subsubsection{Direct and iterative estimation}

The estimation procedure for computing Shapley values is controlled by the \code{iterative} argument in \code{explain()}. 
The iterative estimation procedure with convergence detection described in Section~\ref{sec:iterative} is mainly useful for reducing the runtime of the Shapley values computation when there are many features/feature-groups. 
When there are fewer Shapley values to compute, the computational gain of stopping early is often negated by the cost of computing the bootstrapped standard deviations of the Shapley values.
Therefore, in \pkg{shapr}, the default behavior (\code{iterative = NULL}) is to use the direct estimation procedure when there are five or fewer features (or feature-groups), and the iterative procedure otherwise.
In the former case \code{explain()} also uses all $2^M$ coalitions by default.
The number of coalitions to use can be restricted through the argument \code{max_n_coalitions}. 
For the direct estimation procedure, this is the actual number of unique coalitions to use. The iterative procedure stops when the number of coalitions reaches this number.
Specifics related to the iterative procedure and convergence criterion are set through the \code{iterative_args} argument, where the user can set the convergence criterion parameter $t$ in \eqref{eq:convcrit}, the number of ``burn-in'' coalitions used in the first iteration, 
the maximum number of iterations, etc. For details about defaults and arguments, see the help file of \code{get_iterative_args_default}.
In Section \ref{sec:conditional_shapley_values:examples}, we provide examples with both the direct and iterative estimation procedure.


\subsubsection{Continued estimation}

While the iterative estimation procedure with convergence detection aims to balance runtime and accuracy, there may be cases where further reduction of the estimating uncertainty is desired.
\pkg{shapr} allows continuation of iterative or non-iterative Shapley value computations via the \code{prev_shapr_object} argument in \code{explain()}, which accepts either a \code{(shapr, list)} object or a file path to saved intermediate results from a previous call to \code{explain()}.
For an object \code{ex} of class \code{(shapr, list)}, the path is accessible via \code{ex\$saving_path}, which defaults to a file in \code{tempdir()}.
This feature is useful for resuming interrupted computations or improving the accuracy by continuing iterative estimation. 
To include additional coalitions, adjustments to \code{max_n_coalitions} or the \code{iterative_args} parameters described above may be necessary.
We provide an example of this functionality in Section \ref{sec:conditional_shapley_values:examples}.

\subsubsection{Parallelized batch computations}

Estimation of the conditional expectations in $v(\mathcal{S}) = \E[f(\boldsymbol{x})|\boldsymbol{x}_{\mathcal{S}}] $ is the most computationally intensive part of the Shapley value computation. 
When relying on Monte Carlo-based estimation approaches, these computations may also involve processing a large amount of data and may, therefore, be somewhat memory-intensive  if handled all at once.
To reduce memory usage, \pkg{shapr} supports batch-wise estimation of $v(\mathcal{S})$, discarding intermediate objects and data between batches.
In the iterative estimation procedure, batch computations are performed within each iteration.
To reduce runtime, \pkg{shapr} also supports parallelization over these batches (within every iteration).
The parallelized computations are handled by the \pkg{future} framework \citep{RJ-2021-048}, enabling flexible parallel computations on all major operating systems, in addition to computing clusters, completely controlled by the user outside of \pkg{shapr}. 
We give an example of how to specify the parallelization setup in Section~\ref{sec:conditional_shapley_values:examples}.
Batch computations do increase the runtime by a small amount, and it is therefore advised not to use too many batches if memory consumption is not an issue.
Users can control the batch size and number of them by specifying \code{min_n_batches}, \code{max_batch_size} as named list elements passed to the \code{extra_computation_args} argument of \code{explain()}. If not specified, both default to 10, typically providing a decent trade-off between computation speed and memory consumption.
While parallelization speeds up computations, it also increases memory consumption and incurs some overhead from duplicating data across processes, so it is often beneficial to limit the number of parallel sessions.

\subsubsection{Verbosity and progress reports}

Since conditional Shapley value computations often take time, it can be useful for the user to get feedback on the current stage of the computation process. 
In \pkg{shapr}, the verbosity of the output is controlled by the \code{verbose} argument, taking one or more strings as input. 
It can provide information about the computation to be performed (\code{"basic"}, default), 
the current stage of the estimation procedure (\code{"progress"}), 
how close to convergence the iterative procedure is (\code{"convergence"}), 
intermediate Shapley values estimates, (\code{"shapley"}), and details about
the $v(\mathcal{S})$ estimation process (\code{"vS_details"}).
The output is neatly formatted using the \pkg{cli} package \citep{cli}.

In addition, progress bars for the $v(\mathcal{S})$ computations are available through the \pkg{progressr} package \citep{progressr}, which are specified separately from the \pkg{shapr} package. 
\pkg{progressr} integrates seamlessly with the \pkg{future} parallelization framework \citep{RJ-2021-048} and is, to our knowledge, the only tool that provides progress updates for parallel tasks in \proglang{R}. 
These progress bars can be used alongside or independently of the \code{verbose} argument.

\subsubsection{Comparing approaches}
Following a comparative study of different approaches for estimating $v(\mathcal{S})$, \citet{olsen2024comparative} provide some rough practical guidelines for how to choose the most appropriate approach. 
As the performance may vary quite a bit depending on the actual feature distribution, we generally recommend comparing the performance of multiple approaches using the $\operatorname{MSE}_{v}$ evaluation criterion described in Section \ref{sec:MSEv}.
In \pkg{shapr}, the $\operatorname{MSE}_{v}$ scores are automatically computed and returned by \code{explain()}.
The metric can be plotted using the \code{plot_MSEv_eval_crit()} function, which takes a list of outputs from \code{explain()} as the main argument.


\subsection{Implementation details} \label{sec:conditional_shapley_values:implementation}

Algorithm \ref{alg:explain} provides an overview of the structure of \code{shapr::explain()}. 
As shown, the function is divided into several smaller components, each responsible for a specific part of the computation. 
These components operate on a shared list named \code{internal}, which serves as both input and output, and stores parameters, settings, and intermediate results throughout the process. 
This modular design also facilitates a relatively simple \pkg{shaprpy} \proglang{Python} implementation (see Section \ref{sec:python}).

\begin{algorithm}[ht!]
\caption{Rough structure of \code{shapr::explain()}}\label{alg:explain}
\begin{algorithmic}
\State converged $\gets $ FALSE
\State \code{internal} $\gets$ \code{list()} \Comment{Creates the \code{internal} list}
\State \code{internal} $\gets$ \code{setup()} \Comment{Checks arguments and sets (default) parameters}
\State \textbf{Call} \code{test_predict_model(internal)} \Comment{Tests the prediction model}
\State \textit{Some additional setup for special cases}
\State \textbf{Call} \code{cli_startup(internal)} \Comment{Prints basic verbosity}
\While{converged = FALSE}
    \State \code{internal} $\gets$ \code{shapley_setup(internal)} \Comment{Samples coalitions}
    \State \code{internal} $\gets$ \code{setup_approach(internal)} \Comment{Prepares the approach(es)}
    \State \code{vS_list} $\gets$ \code{compute_vS(internal)} \Comment{Estimates $v(\mathcal{S})$ (in parallel)}
    \State \code{internal} $\gets$ \code{compute_estimates(vS_list, internal)} \Comment{Computes $\phi_j$, sd$(\phi_j)$}
    \State \code{internal} $\gets$ \code{check_convergence(internal)} \Comment{Checks if convergence is reached}
    \State \textbf{Call} \code{save_results()} \Comment{Saves the intermediate results}
    \State \code{internal} $\gets$ \code{prepare_next_iteration(internal)} \Comment{Prepares next iteration}
    \State \textbf{Call} \code{print_iter(internal)} \Comment{Prints iterative estimates}
    \State \code{converged} $\gets$ from \code{internal} \Comment{Set to TRUE if converged}
\EndWhile
\State \code{output} $\gets$ \code{finalize_explanation(internal)} \Comment{Extracts the computations to return}  
\State \code{output} $\gets$ \code{compute_time(output)} \Comment{Computes and gathers the task specific timing}
\textbf{Return} \code{output}
\end{algorithmic}
\end{algorithm}

Since computing (conditional) Shapley values is computationally demanding, we emphasize efficiency in our implementation.
We rely on the fast and memory-efficient \pkg{data.table} package \citep{Rdatatable} for nearly all internal data operations.
To further enhance performance, we write computationally demanding code in C++, using the \pkg{Rcpp} and \pkg{RcppArmadillo} packages \citep{Rcpp,RcppArmadillo}. 
Finally, the computations in the \code{compute_vS} function can be parallelized, with full user control via the \pkg{future} framework \citep{RJ-2021-048}.

Moreover, we employ various methodological techniques to accelerate the computation of $v(\mathcal{S})$.
For example, in the \code{gaussian} approach, we efficiently reuse an initial set of samples when generating $K$ Monte Carlo samples from
$p(\boldsymbol{x}_{\thickbar{\mathcal{S}}} | \boldsymbol{x}_{\mathcal{S}} = \boldsymbol{x}_{\mathcal{S}}^*)$ 
for the potentially many explicands $\boldsymbol{x}^*$ and subsets $\mathcal{S}$ (within the same batch):
First, we generate $p \times K$ standard normally distributed variables and split them into vectors of size $p$.
Then, for each $\mathcal{S}$, we multiply these vectors by the Cholesky decomposition of their associated conditional covariance matrix in a computationally efficient, vectorized manner.
Finally, we add the conditional means, which differ for each explicand $\boldsymbol{x}^*$.
Internal benchmarks have shown that our implementation of this procedure in C++
provides speedups of several orders of magnitude compared to the standard approach of separate sampling solely in \proglang{R}.
This same technique is applied to the internal Gaussian samples in the \code{copula} approach.

\subsection{Examples} \label{sec:conditional_shapley_values:examples}


Below, we provide some basic use cases of the \pkg{shapr} package. 
More extensive examples\linebreak are available through the package's ``General usage'' vignette available through \linebreak\code{vignette("general_usage","shapr")} or the \href{https://norskregnesentral.github.io/shapr/articles/general_usage.html}{online documentation}.

We demonstrate \pkg{shapr} on the \href{https://archive.ics.uci.edu/dataset/275/}{bike sharing} dataset from the UCI Machine Learning Repository. 
We follow \citet{heskes2020causal} who use this dataset and model the daily bike rental counts based on $M=7$ features, where some are hand-crafted from other features in the original dataset. 
The model is a simple \pkg{xgboost} model with default hyperparameters and $100$ trees. 
The features are the number of days since Jan 2011 (\code{trend}), two cyclical variables representing the season (\code{cosyear}, \code{sinyear}), temperature (\code{temp}), temperature feel (\code{atemp}), wind speed (\code{windspeed}), and humidity (\code{hum}).
$80\%$ of the 731 observations are used to train the model, while the remaining $20\%$ of the data are held out for testing (and explanation in the present case).
The script used to prepare the data is available in the \pkg{shapr}
\href{https://github.com/NorskRegnesentral/shapr/tree/master/inst/code_paper}{GitHub repository}.

We first load the required packages and read in the processed data and fitted model:
\begin{CodeChunk}
\begin{CodeInput}
R> library(xgboost)
R> library(data.table)
R> library(shapr)
R> x_explain <- fread(file.path("data_and_models", "x_explain.csv"))
R> x_train <- fread(file.path("data_and_models", "x_train.csv"))
R> y_train <- unlist(fread(file.path("data_and_models", "y_train.csv")))
R> model <- readRDS(file.path("data_and_models", "model.rds"))
\end{CodeInput}
\end{CodeChunk}

We utilize the \pkg{future} and \pkg{progressr} packages to enable parallelized computations (using 4 parallel sessions) with progress updates for the $v(\mathcal{S})$ estimation as follows:
\begin{CodeChunk}
\begin{CodeInput}
R> library(future)
R> library(progressr)
R> future::plan(multisession, workers = 4)
R> progressr::handlers(global = TRUE)
\end{CodeInput}
\end{CodeChunk}

We then show how to explain the 146 explicands in \code{x_explain} with the \code{independence} and \code{ctree} approaches, using a maximum of 40 coalitions:
\begin{CodeChunk}
\begin{CodeInput}
R> # 40 indep
R> exp_40_indep <- explain(model = model,
+                          x_explain = x_explain,
+                          x_train = x_train,
+                          max_n_coalitions = 40,
+                          approach = "independence",
+                          phi0 = mean(y_train),
+                          verbose = NULL,
+                          seed = 1)

R> # 40 ctree
R> exp_40_ctree <- explain(model = model,
+                          x_explain = x_explain,
+                          x_train = x_train,
+                          max_n_coalitions = 40,
+                          approach = "ctree",
+                          phi0 = mean(y_train),
+                          verbose = NULL,
+                          ctree.sample = FALSE,
+                          seed = 1)
\end{CodeInput}
\end{CodeChunk}
We compare the two approaches based on their resulting $\operatorname{MSE}_{v}$:
\begin{CodeChunk}
\begin{CodeOutput}
R> print(exp_40_indep, what = "MSEv")
       MSEv MSEv_sd
      <num>   <num>
 1: 1520051  100972
R> print(exp_40_ctree, what = "MSEv")
       MSEv MSEv_sd
      <num>   <num>
 1: 1169051   70910
\end{CodeOutput}
\end{CodeChunk}
The \code{ctree} method is clearly the best. We thus print its estimated Shapley values.
\newpage

\begin{figure}[ht!]
\centering
\includegraphics[width=1\textwidth]{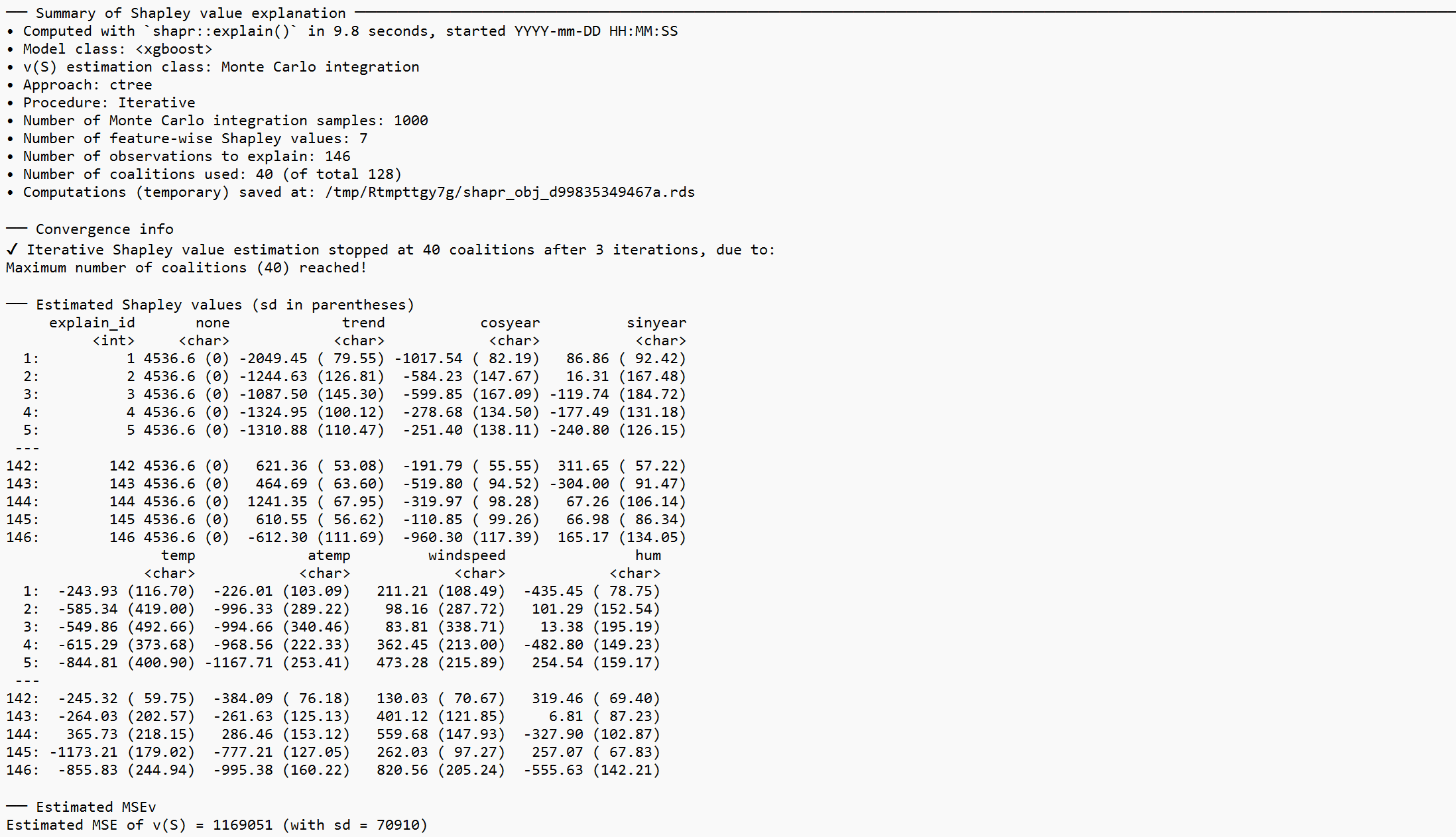}
    \caption{The output displayed in the console when calling \code{summary(exp\_40\_ctree)}.}
    \label{fig:summary_shapr}
\end{figure}

\begin{CodeChunk}
\begin{CodeOutput}
R> print(exp_40_ctree)
      explain_id  none trend cosyear sinyear  temp atemp windspeed     hum
           <int> <num> <num>   <num>   <num> <num> <num>     <num>   <num>
   1:          1  4537 -2049   -1018    86.9  -244  -226     211.2 -435.45
   2:          2  4537 -1245    -584    16.3  -585  -996      98.2  101.29
   3:          3  4537 -1088    -600  -119.7  -550  -995      83.8   13.38
   4:          4  4537 -1325    -279  -177.5  -615  -969     362.4 -482.80
   5:          5  4537 -1311    -251  -240.8  -845 -1168     473.3  254.54
  ---                                                                     
 142:        142  4537   621    -192   311.6  -245  -384     130.0  319.46
 143:        143  4537   465    -520  -304.0  -264  -262     401.1    6.81
 144:        144  4537  1241    -320    67.3   366   286     559.7 -327.90
 145:        145  4537   611    -111    67.0 -1173  -777     262.0  257.07
 146:        146  4537  -612    -960   165.2  -856  -995     820.6 -555.63
\end{CodeOutput}
\end{CodeChunk}
By visually inspecting this subset of explanations, it seems that \code{trend} and \code{temp}/\code{atemp} are highly influential features. Calling \code{summary(exp_40_ctree)} outputs a structured summary of the Shapley value computation to the console, as illustrated in Figure \ref{fig:summary_shapr}.

\begin{figure}[ht]
\centering
         \includegraphics[width=1\textwidth]{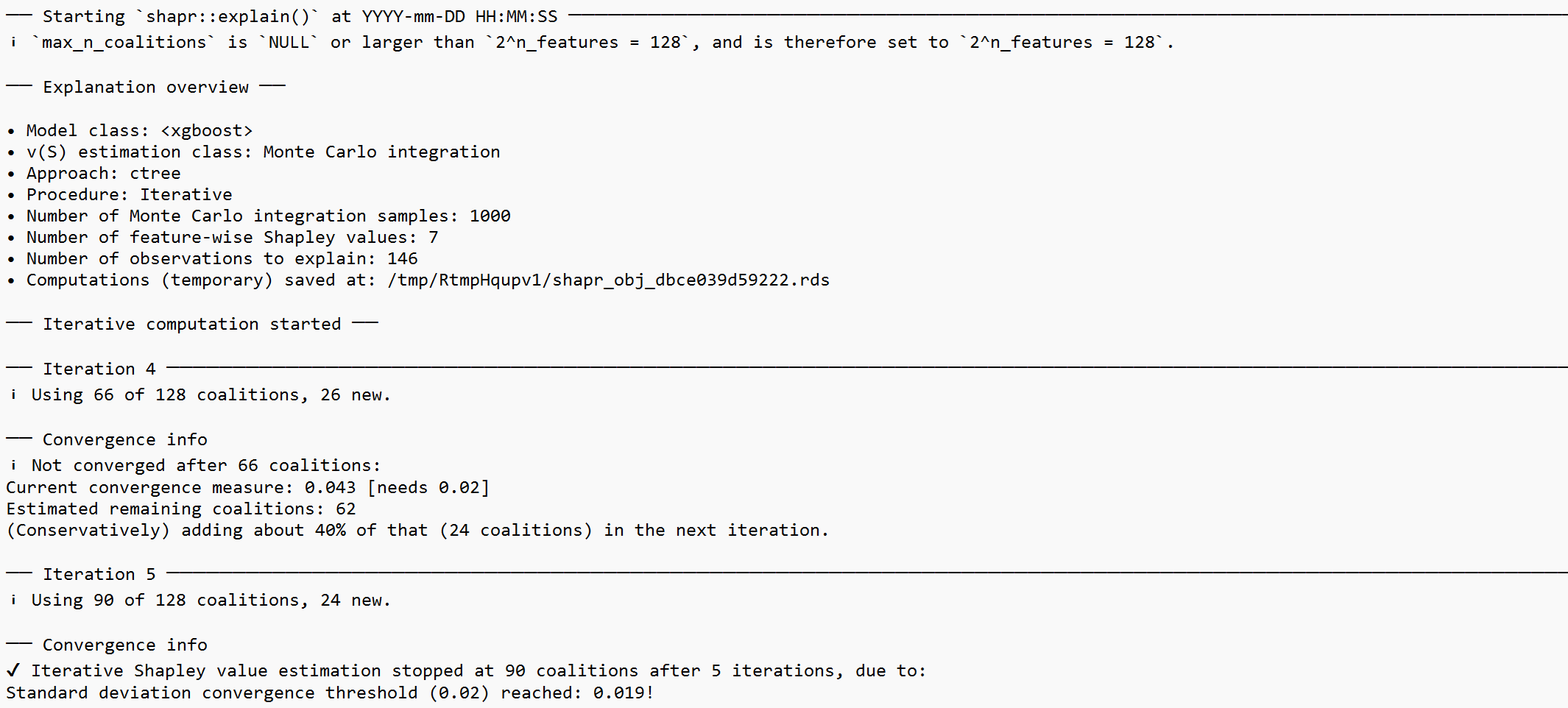}
    \caption{The output to the console when creating \code{exp\_iter\_ctree} with \code{explain()} using \code{verbose = c("basic", "convergence")}.}
    \label{fig:verbose_shapr}
\end{figure}

Next, we show how to resume estimation with the \code{ctree} approach from the first 40 coalitions, and also enable verbose output to monitor convergence. The output is shown in Figure \ref{fig:verbose_shapr}.

\begin{CodeChunk}
\begin{CodeInput}
R> exp_iter_ctree <- explain(model = model,
+                            x_explain = x_explain,
+                            x_train = x_train,
+                            approach = "ctree",
+                            phi0 = mean(y_train),
+                            prev_shapr_object = exp_40_ctree,
+                            ctree.sample = FALSE,
+                            verbose = c("basic", "convergence"),
+                            seed = 1)
\end{CodeInput}
\end{CodeChunk}

\begin{figure}[ht!]
\centering
         \includegraphics[width=0.75\textwidth]{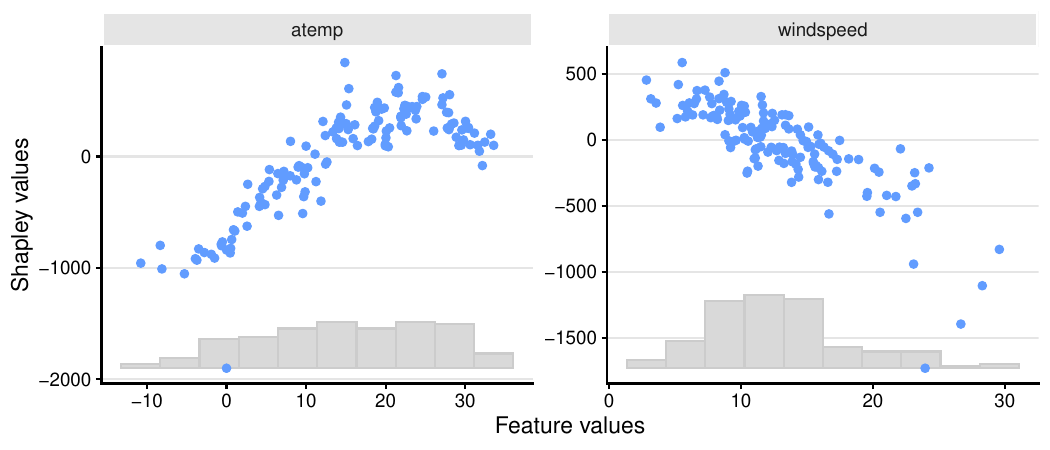}
    \caption{Scatter plot of the Shapley values for the two features \code{atemp} and \code{windspeed}, created by \code{plot.shapr}. The Shapley values are largest for \code{atemp} around 20 and \code{windspeed} less than 10, meaning that in these situations, the temperature and wind speed increases the predicted number of bike rentals the most. For very low \code{atemp} and high \code{windspeed}, the predicted number of bike rentals is greatly reduced, which makes intuitive sense.}
    \label{fig:scatter_ctree}
\end{figure}

To further investigate the results, we create a scatter plot of two of the features (\code{atemp} and \code{windspeed}). The plot is displayed in Figure \ref{fig:scatter_ctree}.
\begin{CodeChunk}
\begin{CodeInput}
R> library(ggplot2)
R> plot(exp_iter_ctree, 
+       plot_type = "scatter", 
+       scatter_features = c("atemp", "windspeed"))
\end{CodeInput}
\end{CodeChunk}

Finally, we showcase how group-wise Shapley value explanations can be provided, using the \code{regression_separate} approach with both a default and tuned \pkg{xgboost} \citep{xgboost} model. 
The features are divided into three groups based on their type (temperature, time and weather).
\begin{CodeChunk}
\begin{CodeInput}
R> group <- list(temp = c("temp", "atemp"),
+                time = c("trend", "cosyear", "sinyear"),
+                weather = c("hum","windspeed"))

R> exp_g_reg <- explain(model = model,
+                       x_explain = x_explain,
+                       x_train = x_train,
+                       phi0 = mean(y_train),
+                       group = group,
+                       approach = "regression_separate",
+                       regression.model = parsnip::boost_tree(
+                         engine = "xgboost", 
+                         mode = "regression"
+                       ),
+                       verbose = NULL,
+                       seed = 1)
\end{CodeInput}
\end{CodeChunk}
\begin{CodeChunk}
\begin{CodeInput}
R> tree_vals <- c(10, 15, 25, 50, 100, 500)
R> exp_g_reg_tuned <- explain(model = model,
+                             x_explain = x_explain,
+                             x_train = x_train,
+                             phi0 = mean(y_train),
+                             group = group,
+                             approach = "regression_separate",
+                             regression.model = parsnip::boost_tree(
+                               trees = hardhat::tune(),
+                               engine = "xgboost", 
+                               mode = "regression"
+                             ),
+                             regression.tune_values = expand.grid(
+                               trees = tree_vals
+                             ),
+                             regression.vfold_cv_para = list(v = 5),
+                             verbose = NULL,
+                             seed = 1)
\end{CodeInput}
\end{CodeChunk}
We then print and compare their resulting $\operatorname{MSE}_{v}$ scores and computation time. 
\begin{CodeChunk}
\begin{CodeOutput}
R> print(exp_g_reg, what = "MSEv")
       MSEv MSEv_sd
      <num>   <num>
 1: 1392003   93288
R> print(exp_g_reg_tuned, what = "MSEv")
       MSEv MSEv_sd
      <num>   <num>
 1: 1367885   92316

R> print(exp_g_reg, what = "timing_summary")
              init_time            end_time total_time_secs total_time_str
                 <POSc>              <POSc>           <num>         <char>
 1: YYYY-MM-DD 09:57:46 YYYY-MM-DD 09:57:48            2.55    2.6 seconds

R> print(exp_g_reg_tuned, what = "timing_summary")
              init_time            end_time total_time_secs total_time_str
                 <POSc>              <POSc>           <num>         <char>
 1: YYYY-MM-DD 09:57:48 YYYY-MM-DD 09:57:55            6.92    6.9 seconds
\end{CodeOutput}
\end{CodeChunk}

\begin{figure}[ht!]
\centering
         \includegraphics[width=0.8\textwidth]{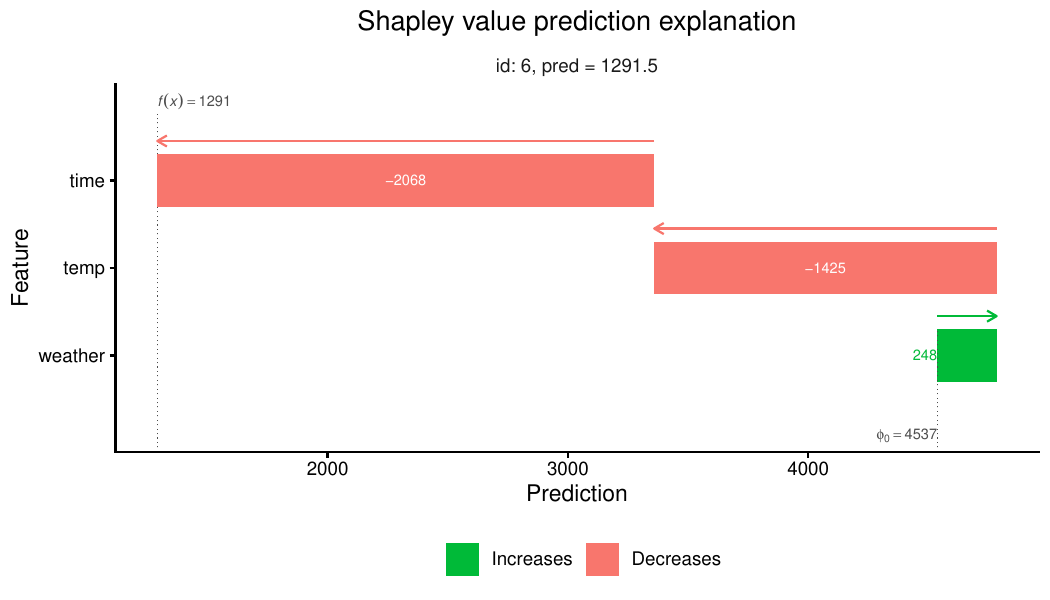}
    \caption{Illustration of a waterfall plot of a single explicand with group-wise Shapley values created by \code{plot.shapr}. 
    The waterfall plot illustrates how the cumulative contributions of feature groups sum to the prediction, starting from the mean prediction/response $\phi_0$ and adding contributions in order of increasing absolute value. For this explicand, the \code{weather} group increases the predicted number of bike rentals slightly, while the \code{time} and \code{temp} groups reduce it considerably.}
    \label{fig:waterfall_group}
\end{figure}

As we can see, the tuned version is marginally better (though far from significant), and it also takes almost three times as long to run.
Finally, we plot group-wise Shapley values for a specific observation (observation 6 in \code{x_explain}) of the tuned version, giving the plot in Figure \ref{fig:waterfall_group}.

\begin{CodeChunk}
\begin{CodeOutput}
R> plot(exp_g_reg_tuned, 
+       index_x_explain = 6, 
+       plot_type = "waterfall")
\end{CodeOutput}
\end{CodeChunk}

\section[Asymmetric and causal Shapley values with shapr]{Asymmetric and causal Shapley values with \pkg{shapr}}
\label{sec:asymmetric_and_causal_shapley_values}

In some cases, we have knowledge about the causal structure within the data we are modeling, but traditional Shapley values do not account for this. Asymmetric Shapley values and causal Shapley values address this limitation by incorporating causal structures into their explanations, thereby offering more informative explanations in these scenarios. 

\subsection{Overview}
\begin{figure}[h]
     \centering
     \hfill
     \begin{subfigure}[b]{0.455\textwidth}
         \centering
         \includegraphics[width=0.7\textwidth]{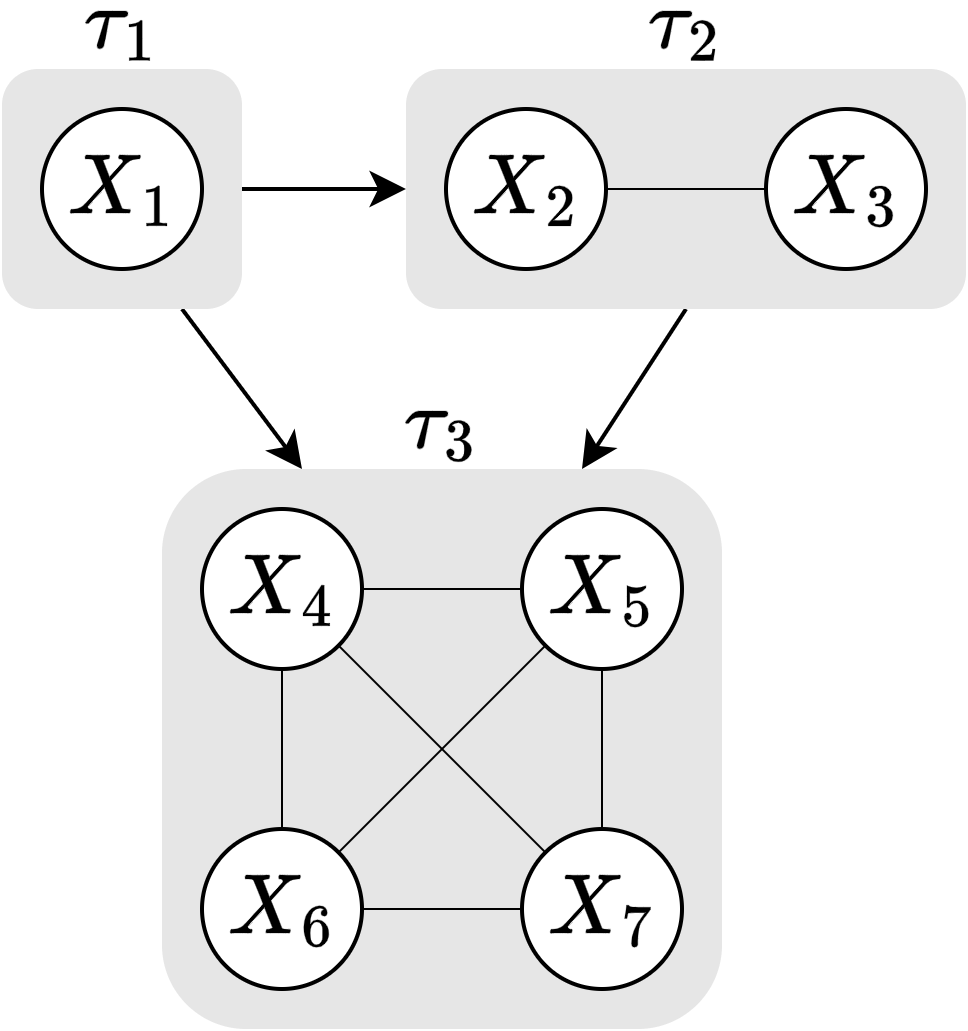}
         \caption{\code{confounding = NULL}}
         \label{fig:causal_ordering_without_confounding}
     \end{subfigure}
     \hfill
     \begin{subfigure}[b]{0.455\textwidth}
         \centering
         \includegraphics[width=0.7\textwidth]{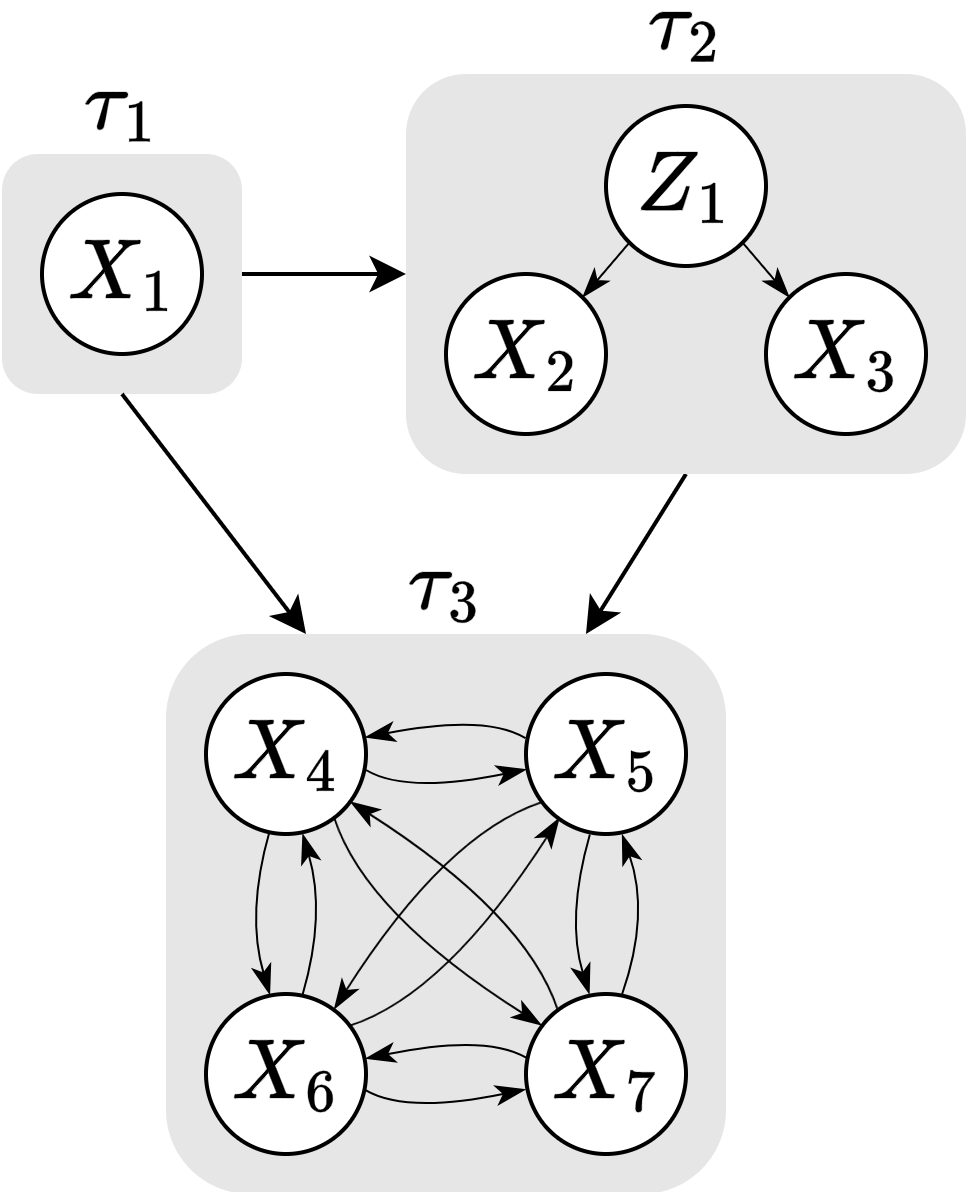}
         \caption{\code{confounding = c(FALSE, TRUE, FALSE)}}
         \label{fig:causal_ordering_with_confounding}
     \end{subfigure}
     \hfill
    \caption{Schematic overview of \code{causal\_ordering = list($\tau_1$ = 1, $\tau_2$ = c(2,3), $\tau_3$ = c(4,5,6,7))} without and with confounding specified in the second component.}
    \label{fig:causal_ordering}
\end{figure}

Asymmetric \citep{frye2020asymmetric} and causal \citep{heskes2020causal} Shapley values aim to incorporate causal knowledge into the explanations. Both frameworks rely on the user specifying a (partial) causal ordering, where features that are treated on an equal footing are linked together with undirected edges and become part of the same chain component $\tau_i$. Edges between chain components $\tau_i$ and $\tau_j$ are directed and represent causal relationships. Additionally, in the causal Shapley value framework, the user must specify if component $\tau_i$ is subject to confounding or not. Together, they form a causal chain graph with directed and undirected edges. Let the chain components with and without confounding be denoted by $\mathcal{T}_{\text{confounding}}$ and $\mathcal{T}_{\,\overline{\text{confounding}}}$, respectively. This allows us to correctly distinguish between dependencies that are due to confounding and mutual interactions. In Figure~\ref{fig:causal_ordering}, we visualize a causal ordering and the corresponding causal chain graph in an $M = 7$-dimensional setting when we assume confounding in $\tau_2$ but no confounding in $\tau_1$ and $\tau_3$. This causal structure is assumed in the example below. 

\citet{frye2020asymmetric} achieve the incorporation of causal knowledge into the explanations by estimating the Shapley values solely based on coalitions that respect the causal ordering. This means that not all $2^M = 128$ coalitions are used, where $M$ is the number of features. For example, in Figure~\ref{fig:causal_ordering}, we see that $X_1$ is the ancestor of $X_2$; thus, asymmetric Shapley values omit coalitions where $X_2$ is included and $X_1$ is excluded. For the causal ordering in Figure~\ref{fig:causal_ordering}, there are $20$ valid coalitions. Only using these will skew the explanations towards distal/root causes \citep[Sec.~3.2]{frye2020asymmetric}.

\citet{heskes2020causal} argue that causal Shapley values offer a more direct and robust way to incorporate causal knowledge than asymmetric Shapley values. They redefine $v(\s)$ to
\begin{align*}
    v(\s) 
   &=
   \E\left[ f(\xsb, \xs) \,|\, \doop(\xs = \xss) \right] 
    = 
    \int f(\xsb, \xss) p(\xsb \,|\, \doop(\xs = \xss)) \diff \xsb,
\end{align*}
where the $\doop$ operator stems from Pearl’s do-calculus \citep{pearl1995causal,pearl2012calculus}. Furthermore, 
\begin{align}
\label{eq:ContributionFunc:do}
\begin{split}
    p(\xsb \,|\, \doop(\xs = \xss))
    = &
    \prod_{\tau \in \mathcal{T}_{\,\text{confounding}}}
    p(\x_{\tau \cap \thickbar{\mathcal{S}}} \mid \x_{\text{pa}(\tau) \cap \thickbar{\mathcal{S}}}, \x_{\text{pa}(\tau) \cap \mathcal{S}}) \times \\
    & \quad 
    \prod_{\tau \in \mathcal{T}_{\,\overline{\text{confounding}}}}
    p(\x_{\tau \cap \thickbar{\mathcal{S}}} \mid \x_{\text{pa}(\tau) \cap \thickbar{\mathcal{S}}}, \x_{\text{pa}(\tau) \cap \mathcal{S}}, \x_{\tau \cap \mathcal{S}}),
\end{split}
\end{align}
where $\text{pa}(\tau)$ are the parent features of chain component $\tau$. For specific causal chain graphs, (\ref{eq:ContributionFunc:do}) simplifies to the marginal $p(\xsb)$ and conditional $p(\xsb \,|\, \xs = \xss)$ distributions. Consequently, the causal Shapley values generalize the marginal and conditional Shapley values \citep[Suppl.\ Cor.\ 1-3]{heskes2020causal}. By considering only the coalitions that respect the causal ordering, we obtain \textit{asymmetric causal Shapley values}. 

Causal Shapley values are computed by altering the Monte Carlo sampling procedure discussed in Section~\ref{sec:background:paradigms} based on the causal chain graph. Specifically, generating $\boldsymbol{x}_{\thickbar{\mathcal{S}}}^{(k)} \sim p(\xsb \,|\, \doop(\xs = \xss))$ consists of a chain of sampling steps defined by the causal ordering. 

\begin{table}[t]
\centering
\centerline{
\begin{adjustbox}{width=1\textwidth}
\begin{tabular}{llllll}
\toprule
Framework         & Sampling             & \code{asymmetric} & \code{causal\textunderscore ordering}  & \code{confounding} & \code{approach} \\ 
\specialrule{.4pt}{2pt}{0pt}
\rowcolor{gray!15} Sym. conditional  & $P(\Xsb|\Xs = \xs)$        & \code{FALSE} & \code{NULL}      & \code{NULL}   & All \\
Asym. conditional & $P(\Xsb|\Xs = \xs)$        & \code{TRUE}  & \code{list(...)} & \code{NULL}   & All \\
\rowcolor{gray!15} Sym. causal       & $P(\Xsb|\doop(\Xs = \xs))$ & \code{FALSE} & \code{list(...)} & \code{c(...)} & All MC-based \\
Asym. causal      & $P(\Xsb|\doop(\Xs = \xs))$ & \code{TRUE}  & \code{list(...)} & \code{c(...)} & All MC-based  \\
\rowcolor{gray!15} Sym. marginal     & $P(\Xsb)$                  & \code{FALSE} & \code{NULL}      & \code{TRUE}   & \code{indep.}, \code{gaussian} \\
\specialrule{.8pt}{0pt}{2pt}
\end{tabular}
\end{adjustbox}
}
\caption{\small Overview of the Shapley value methodologies supported by \pkg{shapr} and how to compute them by altering the \code{asymmetric}, \code{causal\textunderscore ordering}, and \code{confounding} arguments in the \code{explain()} function. Setting \code{causal\textunderscore ordering} to \code{NULL} is equivalent to a causal ordering with one component containing all (groups of) features. By \code{list(...)}, we mean a list of vectors indicating the (groups of) features in each component in the partial ordering, and \code{c(...)} represents a vector of booleans. Finally, \code{approach} indicates the estimation approaches in Section~\ref{sec:background:approaches} compatible with each framework.}
\label{tab:overview_SV}
\end{table}

\subsection{Implementation details}
The \texttt{shapr} package implements both asymmetric and causal Shapley values by adjusting the \code{asymmetric}, \code{causal_ordering}, and \code{confounding} arguments in the \code{explain()} function (Section~\ref{shapr:R:basic_usage}). Asymmetric Shapley values are partially implemented in \pkg{shapFlex} \citep{redell2019shapley}, though it is currently in an experimental state and has not been maintained for the past 5 years. Causal Shapley values are implemented in \pkg{CauSHAPley} \citep{heskes2020causal}, building on an older version of \pkg{shapr} and is limited to the \code{gaussian} approach only.

In Table~\ref{tab:overview_SV}, we provide an overview of how to compute the different Shapley value versions with \pkg{shapr}. 
Asymmetric Shapley values are compatible with all estimation approaches discussed in Section~\ref{sec:background:approaches}, whereas causal Shapley values are only applicable to the Monte Carlo-based approaches. Both methodologies support groups of features, provided the causal ordering is specified at the group level rather than for individual features. All approaches estimate the marginal distributions by sampling from the training data, except the \code{gaussian} method, which samples from the marginals of the estimated Gaussian distribution.

Asymmetric Shapley values are implemented by computing the allowed coalitions based on the causal graph and either sampling from these coalitions or using all of them. To compute causal Shapley values, \pkg{shapr} computes and iterates over the steps in the sampling chains needed to generate $\boldsymbol{x}_{\thickbar{\mathcal{S}}}^{(k)} \sim p(\xsb \,|\, \doop(\xs = \xss))$ for each coalition separately. The separate treatment ensures that all other functionalities described in Section~\ref{sec:package_functinality} for conditional Shapley values also apply to causal Shapley values.

A drawback of this separate treatment is that some sampling steps can occur in multiple chains, leading to repeated estimation of the corresponding conditional distributions. This has no significant runtime impact for easily trained approaches like \code{gaussian} and \code{copula}, but it can affect slower approaches like \code{ctree}. This issue does not influence the \code{vaeac} method, as it identifies the unique sampling steps across all chains and trains on these once. Additionally, step-wise sampling requires all but the first step in a chain to call the approach $K$ times per explicand, where $K$ is the number of Monte Carlo samples. This increased number of calls will impact the runtime of \code{explain()} for slower approaches. 
For computing asymmetric Shapley values in medium to high dimensions or when runtime is a concern, we therefore recommend using the optimized \code{gaussian} and \code{copula} approaches when applicable and the \code{vaeac} approach otherwise.

\subsection{Example}
Below we demonstrate the asymmetric and causal Shapley value frameworks on the same data and model as in Section~\ref{sec:conditional_shapley_values:examples}. 
More extensive examples are available through the package's vignette ``Asymmetric and causal Shapley value explanations'' available through \newline
\code{vignette("asymmetric_causal", "shapr")} or the \href{https://norskregnesentral.github.io/shapr/asymmetric_causal.html}{online documentation}.
Once again, we follow \citet{heskes2020causal} who consider the first three features (\code{trend}, \code{cosyear}, and \code{sinyear}) to be potential causes of the four weather-related features (\code{temp}, \code{atemp}, \code{windspeed}, \code{hum}). Following \citet{heskes2020causal}, we set $\tau_1 = \set{\text{\code{trend}}}$, $\tau_2 = \set{\text{\code{cosyear}, \code{sinyear}}}$, and $\tau_3 = \set{\text{\code{temp}, \code{atemp}, \code{windspeed}, \code{hum}}}$, assuming confounding only in $\tau_2$. This setup corresponds to the causal chain graph in Figure~\ref{fig:causal_ordering}.

\begin{figure}
    \centering
    \includegraphics[width=1\linewidth]{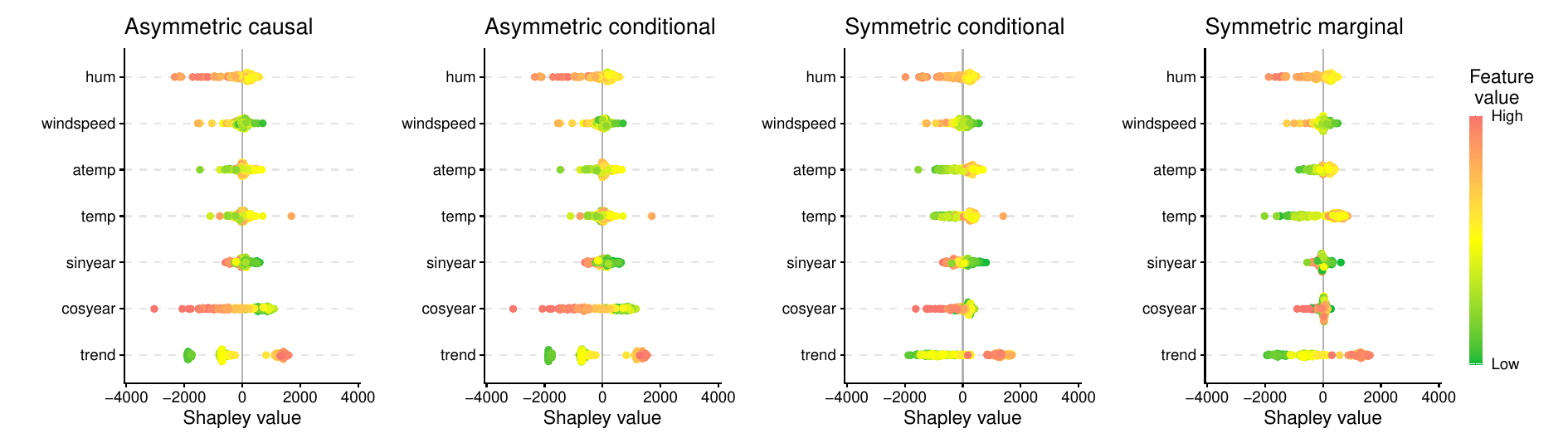}
    \caption{Beeswarm plots of the Shapley values computed by different Shapley value versions.}
    \label{fig:BeeSwarm_Shapley}
\end{figure}

We compute and compare four different Shapley value variants: ``Asymmetric causal'', ``Asymmetric conditional'', ``Symmetric conditional'' and ``Symmetric marginal'' in accordance with the descriptions in Table \ref{tab:overview_SV}, all assuming a Gaussian feature distribution. 
We begin by specifying the causal and confounding structure outlined above.
\begin{CodeChunk}
\begin{CodeInput}
R> causal_order0 <- list("trend",
+                        c("cosyear", "sinyear"),
+                        c("temp", "atemp", "windspeed", "hum"))

R> confounding0 <- c(FALSE, TRUE, FALSE)
\end{CodeInput}
\end{CodeChunk}
Next, we define the four variants, compute their Shapley values, and generate a list of so-called beeswarm plots to compare the results.
\begin{CodeChunk}
\begin{CodeInput}
R> exp_names <- c("Asymmetric causal", "Asymmetric conditional",
+                 "Symmetric conditional", "Symmetric marginal")

R> causal_ordering_list <- list(causal_order0, causal_order0, NULL, NULL)
R> confounding_list <- list(confounding0, NULL, NULL, TRUE)
R> asymmetric_list <- list(TRUE, TRUE, FALSE, FALSE)

R> plot_list <- list()
R> for(i in seq_along(exp_names)){
+    exp_tmp <- explain(model = model,
+                       x_train = x_train,
+                       x_explain = x_explain,
+                       approach = "gaussian",
+                       phi0 = mean(y_train),
+                       asymmetric = asymmetric_list[[i]],
+                       causal_ordering = causal_ordering_list[[i]],
+                       confounding = confounding_list[[i]],
+                       seed = 1,
+                       verbose = NULL)
+ 
+    plot_list[[i]] <- plot(exp_tmp, 
                            plot_type = "beeswarm", 
                            print_ggplot = FALSE) +
+      ggplot2::ggtitle(exp_names[i]) + ggplot2::ylim(-3700, 3700)
+ }
\end{CodeInput}
\end{CodeChunk}
Finally, we use the \pkg{patchwork} package \citep{patchwork} to combine the beeswarm plots into a single figure, as shown in  Figure~\ref{fig:BeeSwarm_Shapley}.
\begin{CodeChunk}
\begin{CodeInput}
R> library(patchwork)
R> patchwork::wrap_plots(plot_list, nrow = 1) + 
+    patchwork::plot_layout(guides = "collect")
\end{CodeInput}
\end{CodeChunk}

\citet{heskes2020causal} argue that asymmetric Shapley values focus on root causes, symmetric marginal Shapley values emphasize direct effects, and symmetric causal Shapley values consider both for a more comprehensive explanation. 
As seen from the beeswarm plots in Figure \ref{fig:BeeSwarm_Shapley}, there is only a minimal difference between the causal and conditional asymmetric methods in this case.
The asymmetric frameworks often assign larger Shapley values (in magnitude) to the root cause \code{cosyear} (season) compared to the direct effect of \code{temp} (temperature), whose Shapley values are near zero. 
Conversely, the two symmetric frameworks attribute more importance to \code{temp}, while clustering \code{cosyear} near zero, especially the marginal version. 
These observations align with \citet{heskes2020causal}.

\section[The shaprpy Python library]{The \pkg{shaprpy} \proglang{Python} library}
\label{sec:python}

While the \pkg{shapr} package provides methods for computing conditional Shapley values for models fitted in \proglang{R}, it naturally leaves models fitted in \proglang{Python} unsupported.\footnote{While packages like \pkg{reticulate} \citep{reticulate} allow calling \proglang{Python} code from within \proglang{R}, it is not straightforward to efficiently load \proglang{Python} \textit{objects} within such sessions.}
The goal of \pkg{shaprpy} is to allow users to compute Shapley values for models fitted in \proglang{Python}, with the ease of doing so directly and efficiently within the \proglang{Python} environment.
\pkg{shaprpy} is a lightweight wrapper around \pkg{shapr}, internally calling functions from the \proglang{R} package.

Below, we introduce the \pkg{shaprpy} library, discuss its usage, supported models, and limitations, and provide an example.

\subsection{The library}

The \pkg{shaprpy} wrapper leans heavily on the \pkg{rpy2} library \citep{gautier2023rpy2}, which provides a neat and function-rich interface for running \proglang{R} embedded in a \proglang{Python} script, function or library.
To use \pkg{shaprpy}, \proglang{R} must be installed, along with \pkg{shapr}, its dependencies and any additional packages required for specific functionality. 
We provide installation instructions under the \textit{Python} pane in our \href{https://norskregnesentral.github.io/shapr/shaprpy.html}{online documentation}.

We chose to implement \pkg{shaprpy} as a wrapper, rather than a full re-implementation in \proglang{Python}, in order to minimize the maintenance required on the \proglang{Python} side.
\pkg{shaprpy} closely follows the structure of \pkg{shapr} as outlined in Algorithm \ref{alg:explain}, and delegates all tasks that do not require direct model access to the existing \proglang{R} functions listed in the algorithm.
For this reason most bug fixes, new methodologies, and additional features in \pkg{shapr} will be directly applicable to \pkg{shaprpy} as well.
Note that \pkg{rpy2} has limited support on Windows. \pkg{shaprpy} has mainly been tested on Linux. 

Similar to \pkg{shapr}, the main function for user interaction in \pkg{shaprpy} is \code{explain()}. 
It takes all the same inputs as the \proglang{R} package, except the \code{prev_shapr_object} object for continued estimation, which is not supported in \pkg{shaprpy}. 
Note also that additional arguments to the different approaches are passed with underscores instead of dots, e.g., \code{ctree_minbucket} instead of \code{ctree.minbucket}. 
The \proglang{Python} version of \code{explain()} supports the core functionality from the \proglang{R} version, including all \textit{approaches} in Section \ref{sec:background:approaches}, group-wise Shapley values, direct and iterative estimation, progress reports, and causal and asymmetric Shapley values. 
It currently does not support the aforementioned continued estimation functionality and parallelization, and there is currently no \code{explain_forecast()} counterpart in \pkg{shaprpy}.
Overall, \pkg{shaprpy} provides stable support for its main features, but it is not yet as extensively developed or tested as \pkg{shapr}, and should therefore be regarded as somewhat less mature.
\pkg{shaprpy} has native support for explaining models fitted with \code{sklearn}, \code{xgboost}, and \code{keras}, but, just like in \pkg{shapr}, custom models can be explained by supplying a suitable prediction function via the \code{predict_model} argument.

The output of \code{shaprpy.explain()} is an object of class \code{Shapr}. The most important methods for \code{Shapr} are \code{print()} and \code{summary()}, which are pure wrappers for the corresponding methods in \proglang{R}, as well as \code{get_results()}, which wraps the \code{get_results()} function from \proglang{R}. 
The standard \proglang{Python} methods \code{__str__()} and \code{__repr__()} mirror the behavior of the \code{print()} method of the \code{Shapr} class, for convenient direct inspection of the computed Shapley values.
We also provide the \code{to_shap()} method which transforms a \code{Shapr} object to the \code{Explanation} class used by the \pkg{shap} \proglang{Python} package \citep{shap}, thereby allowing users to leverage the full suite of plotting functionality of that package. Some basic usage is illustrated below.


\subsection{Examples}

Since \pkg{shaprpy} is a wrapper for \pkg{shapr} with essentially the same functionality, we limit ourselves to a simple example showcasing that we can reproduce the result for the initial \code{ctree} approach of Section \ref{sec:conditional_shapley_values:examples} in \proglang{Python}.\footnote{Due to randomness in any sampling-dependent approach, the results of most approaches are not exactly reproducible in \proglang{Python}. However, disabling the node sampling in \code{ctree} with \code{ctree\_sample = False} (\proglang{Python})/ \code{ctree.sample = FALSE} (\proglang{R}), leaves that approach deterministic and, therefore, reproducible.}

We import the required libraries, read in the same data and model used in Section \ref{sec:conditional_shapley_values:examples}.
\begin{CodeChunk}
\begin{CodeInput}
>>> import xgboost as xgb
>>> import pandas as pd
>>> import urllib.request
>>> from shaprpy import explain

>>> # Read data
>>> x_train = pd.read_csv("data_and_models/" + "x_train.csv")
>>> x_explain = pd.read_csv("data_and_models/" + "x_explain.csv")
>>> y_train = pd.read_csv("data_and_models/" + "y_train.csv")

>>> # Load the XGBoost model from the raw format and add feature names
>>> model = xgb.Booster()
>>> model.load_model("data_and_models/" + "model.ubj")
>>> model.feature_names = x_train.columns.tolist() 
\end{CodeInput}
\end{CodeChunk}

We then explain the model using the \code{ctree} approach and restrict the number of coalitions to 40, exactly as done for the \code{exp_40_ctree} call to \proglang{R}'s \code{shapr::explain()} in Section \ref{sec:conditional_shapley_values:examples}.
\begin{CodeChunk}
\begin{CodeInput}
>>> exp_40_ctree = explain(model = model,
...                        x_train = x_train,
...                        x_explain = x_explain,
...                        approach = "ctree",
...                        phi0 = y_train.mean().item(),
...                        max_n_coalitions=40,
...                        ctree_sample = False,
...                        seed = 1)
\end{CodeInput}
\end{CodeChunk}

Printing and comparing the produced Shapley value estimates and the $\operatorname{MSE}_{v}$ scores to those produced in \proglang{R}, we see that they are identical.
\begin{CodeChunk}
\begin{CodeInput}
>>> # Print the Shapley values
>>> exp_40_ctree.print() # Alt: print(exp_40_ctree) (or exp_40_ctree interactively)

\end{CodeInput}
\begin{CodeOutput}
     explain_id  none trend cosyear sinyear  temp atemp windspeed     hum
          <int> <num> <num>   <num>   <num> <num> <num>     <num>   <num>
  1:          1  4537 -2049   -1018    86.9  -244  -226     211.2 -435.45
  2:          2  4537 -1245    -584    16.3  -585  -996      98.2  101.29
  3:          3  4537 -1088    -600  -119.7  -550  -995      83.8   13.38
  4:          4  4537 -1325    -279  -177.5  -615  -969     362.4 -482.80
  5:          5  4537 -1311    -251  -240.8  -845 -1168     473.3  254.54
 ---                                                                     
142:        142  4537   621    -192   311.6  -245  -384     130.0  319.46
143:        143  4537   465    -520  -304.0  -264  -262     401.1    6.81
144:        144  4537  1241    -320    67.3   366   286     559.7 -327.90
145:        145  4537   611    -111    67.0 -1173  -777     262.0  257.07
146:        146  4537  -612    -960   165.2  -856  -995     820.6 -555.63
\end{CodeOutput}
\end{CodeChunk}

\begin{CodeChunk}
\begin{CodeInput}
>>> # Print the MSE of the v(S)
>>> exp_40_ctree.print(what = "MSEv")
\end{CodeInput}
\begin{CodeOutput}
      MSEv MSEv_sd
     <num>   <num>
1: 1169051   70910
\end{CodeOutput}
\end{CodeChunk}

\begin{figure}
    \centering
    \includegraphics[width=1\linewidth]{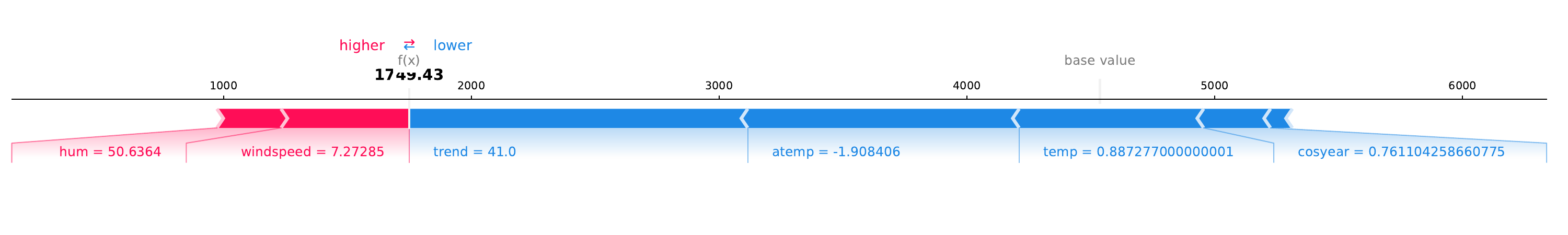}
    \caption{Force plot for the 8th observation in the Python code example, produced by the \pkg{shap} library.}
    \label{fig:Force_plot}
\end{figure}

Moreover, we may produce e.g.~a `force plot' for a single observation (observation 8 in \code{x_explain}) by utilizing the existing plotting functionality of the \pkg{shap} library. The plot is displayed in Figure \ref{fig:Force_plot}. Several other types of visualization are also available; see the full list in the \code{shap} library's \href{https://shap.readthedocs.io/en/latest/api_examples.html#plots}{online documentation}. 

\begin{CodeChunk}
\begin{CodeInput}
>>> from shap import plots as shap_plt
>>> import matplotlib.pyplot as plt

>>> exp_40_ctree_shap = exp_40_ctree.to_shap() # Convert to shap's object class
>>> shap_plt.force(exp_40_ctree_shap[8-1], matplotlib = True) # Display plot
\end{CodeInput}
\end{CodeChunk}

\section{Conditional Shapley values for explaining forecasts with shapr} \label{sec:conditional_shapley_values_forecast}
The \textbf{shapr} package is very flexible in terms of which models can be explained, and the regular \code{explain()} function can also explain time series models. 
However, this can be somewhat cumbersome and inefficient which is why we also provide the specialized function \code{explain_forecast()}.

Given an arbitrary time series model which models a time series $\boldsymbol{y}$ of length $T$, where each observation is written as $y_t, t \in \lbrace 1,...,T\rbrace$, we want to explain a forecast of length $h$. For a simple auto-regressive model, this can be formulated as a regression problem with $h$ outputs and the auto-regressive terms as the inputs that should be explained. However, when introducing multivariate models and exogenous regressors this quickly becomes complicated.

This section outlines the \code{explain\_forecast()} function and how it is used. We also provide some details about how it is implemented, in addition to some brief code examples.

\subsection[The explain forecast() function]{The \code{explain\_forecast()} function}
The \code{explain\_forecast()} function is very similar to its more general \code{explain()} counterpart, with some key differences. Instead of providing the \code{x_explain} and \code{x_train} datasets, \code{explain\_forecast()} takes the following data-related arguments:

\begin{itemize}
    \item \code{y}: One or more endogenous variables used in the model.
    \item \code{xreg}: One or more exogenous variables used in the model. This should also contain observations for the $h$ observations $T + 1,...,T+h$ as they are also imputed when explaining the forecast.
    \item \code{train_idx}: The time indices $\mathcal{T}_{\text{train}}$ which should be used as training data. If this is set to \code{NULL}, it will default to all the available indices not selected in \code{explain_idx}.
    \item \code{explain_idx}: The time indices $\mathcal{T}_{\text{explain}}$ which should be used as starting points for the forecast(s) that should be explained.
    \item \code{explain_y_lags}: A vector containing the number of lags per variable in \code{y} which should be explained.
    \item \code{explain_xreg_lags}: A vector containing the number of lags per variable in \code{xreg} which should be explained.
    \item \code{horizon}: The forecast horizon to be explained.
    \item \code{group\_lags}: A logical denoting if each variable should be explained as a group or if individual lags should be explained independently. The default is \code{TRUE}.
\end{itemize}

The idea of this way of providing the data to the function is to make it more intuitive and less cumbersome for time series forecasting settings. For pure auto-regressive models it is natural to think of a model as a function of $p$ lags which provides an output of length $h$, $f(\boldsymbol{y}_{t - 1}, ..., \boldsymbol{y}_{t - p})$. For moving average models, the formulation is not as obvious, but considering that the MA process will depend more on recent lags than earlier ones, it is possible to truncate the number of lags necessary to make the computation feasible. 
Note that the \textbf{shapr} package makes no assumptions about how many lags are needed to make a prediction when starting from point $t$, it is instead up to the user and the formulation of the function passed as the \code{predict\_model} argument.

Natively, \code{stats::Arima} and \code{stats::ar} are supported model classes. Other model classes can be explained by providing the \code{predict_model} (and optionally \code{get_model_specs}) argument(s). For \code{explain_forecast()}, the \code{predict_model()} function takes additional arguments. The required format is outlined in Appendix \ref{app:predict_model_forecast}.

\subsection{Examples}

Below, we provide a few basic examples of use-cases for \code{explain_forecast()}, once again utilizing the same \href{https://archive.ics.uci.edu/dataset/275/}{bike sharing} dataset.
More extensive examples are available in the Section ``Explaining forecasting models'' in the package's ``General usage'' vignette available through \code{vignette("general_usage","shapr")} or the \href{https://norskregnesentral.github.io/shapr/general_usage.html}{online documentation}.

We start by loading the full dataset (before splitting into a training and explanation set), containing daily ordered observations of the different variables.
\begin{CodeChunk}
\begin{CodeInput}
R> x_full <- fread(file.path("data_and_models", "x_full.csv"))
\end{CodeInput}
\end{CodeChunk}

In the first example, we will build a basic AR(2) model for the variable \textit{temp}, which is the temperature registered on each day. We use the first 729 observations for training the model.
\begin{CodeChunk}
\begin{CodeInput}
R> data_fit <- x_full[seq_len(729), ]
R> model_ar <- ar(data_fit$temp, order = 2)
\end{CodeInput}
\end{CodeChunk}

Now, assume we wish to explain forecasts three steps ahead (\code{horizon = 3}) from the two last time steps (\code{explain_idx = 730:731}), i.e.~explaining the forecasts of $t = 731, 732, 733$ from $t=730$ and $t = 732, 733, 734$ from $t=731$. 
First, we need to choose a reference/baseline value $\phi_0$, representing a prediction with no knowledge.
One option is the last observations before the forecasts, which then gives explanations of the deviation from a naive extrapolation of those values. 
Here, we instead compare against the sample mean of the full time series, and use that reference value for all forecast horizons.
We set \code{explain\_y\_lags = 2} to decompose the forecasts onto the two previous observations, and \code{group_lags = FALSE} to represent their effects separately.
We then compute Shapley value explanations with the \code{empirical} approach as follows:
\newpage
\begin{CodeChunk}
\begin{CodeInput}
R> phi0_ar <- rep(mean(data_fit$temp), 3)
R> exp_fc_ar <- explain_forecast(model = model_ar,
+                                y = x_full[, "temp"],
+                                train_idx = 2:729,
+                                explain_idx = 730:731,
+                                explain_y_lags = 2,
+                                horizon = 3,
+                                approach = "empirical",
+                                phi0 = phi0_ar,
+                                group_lags = FALSE,
+                                seed = 1)
\end{CodeInput}
\end{CodeChunk}

We obtain the result below, where we can see that the first lag has a stronger effect than the second over the full forecast horizon, and for the forecast starting from $t = 731$ the effect is overall more pronounced.
\begin{CodeChunk}
\begin{CodeOutput}
R> print(exp_fc_ar)
    explain_idx horizon  none temp.1 temp.2
          <int>   <int> <num>  <num>  <num>
 1:         730       1  15.3  -5.98  -4.67
 2:         731       1  15.3  -7.82  -4.75
 3:         730       2  15.3  -5.62  -4.37
 4:         731       2  15.3  -7.35  -4.46
 5:         730       3  15.3  -5.28  -4.10
 6:         731       3  15.3  -6.90  -4.18
\end{CodeOutput}
\end{CodeChunk}

Our second example creates an ARIMAX model of order $(2, 0, 0)$ with the variable \textit{windspeed} from the same dataset as an exogenous regressor. 

\begin{CodeChunk}
\begin{CodeInput}
R> model_arimax <- arima(data_fit$temp, 
+                        order = c(2, 0, 0), 
+                        xreg = data_fit$windspeed)
R> phi0_arimax <- rep(mean(data_fit$temp), 2)
\end{CodeInput}
\end{CodeChunk}

Most arguments are equivalent to the previous example, but note here that we also pass the exogenous regressor as \code{xreg}, along with \code{explain_xreg_lags = 1}, meaning that we want to explain a single lag of the exogenous variable \textit{windspeed}. 
For the sake of illustration, we reduce the forecast horizon to two (\code{horizon = 2}) for a single forecast starting point (\code{explain_idx = 729}), and set \code{group_lags = TRUE} to combine the lag effects of, respectively, the endogenous variable (\textit{temp}) and the exogenous regressor (\textit{windspeed}) into single Shapley values.

\begin{CodeChunk}
\begin{CodeInput}
R> exp_fc_arimax <- explain_forecast(model = model_arimax,
+                                    y = data_full[, "temp"],
+                                    xreg = data_full[, "windspeed"],
+                                    train_idx = 2:728,
+                                    explain_idx = 729,
+                                    explain_y_lags = 2,
+                                    explain_xreg_lags = 1,
+                                    horizon = 2,
+                                    approach = "empirical",
+                                    phi0 = phi0_arimax,
+                                    group_lags = TRUE,
+                                    seed = 1)
\end{CodeInput}
\end{CodeChunk}

We see that the specific values of the \textit{windspeed} variable have a small effect on the forecast compared to the auto-regressive effect of \textit{temp} itself, but the effect of \textit{windspeed} is roughly double on the second step of the forecast compared to the first.

\begin{CodeChunk}
\begin{CodeOutput}
R> print(exp_fc_arimax)
    explain_idx horizon  none  temp windspeed
          <int>   <int> <num> <num>     <num>
 1:         729       1  15.3 -8.90     -1.05
 2:         729       2  15.3 -8.59     -2.11
\end{CodeOutput}
\end{CodeChunk}

\section{Summary and discussion} \label{sec:discussion}

The present paper introduces the \pkg{shapr} package, which computes Shapley value explanations for predictive models in \proglang{R}.
The package stands out from other software providing Shapley value-based prediction explanation by focusing on \textit{conditional} Shapley value estimates and implementing a comprehensive list of recently developed methods for estimating these.
Additionally, the package supports computation of causal and asymmetric Shapley values when (partial) causal information is accessible.
Specialized functionality is also provided for explaining forecasts from time series models with multiple forecast horizons.
All of this is packaged with flexible model support, parallelized computations, convergence detection, and visualization tools.
Finally, the accompanying \pkg{shaprpy} \proglang{Python} library makes conditional Shapley value explanations directly accessible in \proglang{Python}, through a wrapper for the core functionality of the \pkg{shapr} \proglang{R} package.
Below, we outline a few potential extensions and enhancements that could be considered for future versions of the software.

While \pkg{shaprpy} ports most functionality to \proglang{Python}, it has some limitations, such as the lack of plotting features. 
Since the popular \pkg{shap} library already has lots of nice plotting functionality for prediction explanations based on Shapley values, creating wrappers for their plotting functionality would be a convenient extension.
Additionally, adapting the \code{explain_forecast()} function from Section \ref{sec:conditional_shapley_values_forecast}, and adding support for parallelization and continued estimation for \proglang{Python}, are natural extensions.

A valuable methodological enhancement to the \pkg{shapr} package, could be to start with several different \textit{approaches} from Section \ref{sec:background:approaches} in a \textit{burn-in} period with a smaller number of coalitions. 
Then, based on the $\operatorname{MSE}_{v}$ metric in Section \ref{sec:MSEv}, we could proceed only with the best-performing estimation approach. 
Implementing this as an automatic step before the iterative estimation procedure could offer a robust and user-friendly solution, which reduces the risk of using approaches inappropriate for the specific data.

SAGE (Shapley Additive Global Explanations) \citep{covert2020understanding} is an explanation method that measures feature importance at the population level rather than for individual predictions.
It decomposes the (expected) training loss across features instead of decomposing individual predictions.
The same justifications regarding marginal and conditional expectations apply also for this case, and the majority of the approaches in Section \ref{sec:background:approaches} can be modified to estimate expected loss instead.
Extending \pkg{shapr} to support SAGE-like decompositions would offer a more comprehensive toolset, enabling both local explanations and global feature importance assessments.




\section*{Acknowledgments}
We thank Didrik Nielsen for creating the initial version of the \pkg{shaprpy} \proglang{Python}
library and all other \href{https://github.com/NorskRegnesentral/shapr/graphs/contributors}{code contributors to \pkg{shapr}}.

The work by Martin Jullum, Lars Henry Berge Olsen and Annabelle Redelmeier has been funded by the Norwegian Research Council's Center for Research-based Innovation BigInsight, project number 237718, and Integreat Center of Excellence, project number 332645 in addition to EU's HORIZON Research and Innovation Programme, project ENFIELD, grant number 101120657.

\bibliography{refs}

@article{covert2020understanding,
  title={{Understanding Global Feature Contributions with Additive Importance Measures}},
  author={Covert, Ian and Lundberg, Scott M and Lee, Su-In},
  journal={Advances in Neural Information Processing Systems},
  volume={33},
  pages={17212--17223},
  year={2020}
}

@Manual{reticulate,
  title = {{\pkg{reticulate}: Interface to '\proglang{Python}'}},
  author = {Kevin Ushey and JJ Allaire and Yuan Tang},
  year = {2024},
  note = {\proglang{R} package version 1.39.0},
  url = {https://rstudio.github.io/reticulate/},
}

@Manual{Rcpp,
    title = {{\pkg{Rcpp}: Seamless \proglang{R} and \proglang{C++} Integration}},
    author = {Dirk Eddelbuettel and Romain Francois and JJ Allaire and
      Kevin Ushey and Qiang Kou and Nathan Russell and Inaki Ucar and
      Douglas Bates and John Chambers},
    year = {2024},
    note = {\proglang{R} package version 1.0.14},
    url = {https://CRAN.R-project.org/package=Rcpp},
  }

@Manual{RcppArmadillo,
    title = {{\pkg{RcppArmadillo}: '\pkg{Rcpp}' Integration for the '\pkg{Armadillo}'
      Templated Linear Algebra Library}},
    author = {Dirk Eddelbuettel and Romain Francois and Doug Bates and
      Binxiang Ni and Conrad Sanderson},
    year = {2024},
    note = {\proglang{R} package version 14.4.2-1},
    url = {https://CRAN.R-project.org/package=RcppArmadillo},
  }

@Manual{Rdatatable,
  title = {{\pkg{data.table}: Extension of `data.frame`}},
  author = {Tyson Barrett and Matt Dowle and Arun Srinivasan and Jan Gorecki and Michael Chirico and Toby Hocking and Benjamin Schwendinger},
  year = {2024},
  note = {\proglang{R} package version 1.17},
  url = {https://r-datatable.com},
}

@article{kokhlikyan2020captum,
  title={{\pkg{Captum}: A Unified and Generic Model Interpretability Library for \pkg{PyTorch}}},
  author={Narine Kokhlikyan and Vivek Miglani and Miguel Martin and Edward Wang and Bilal Alsallakh and Jonathan Reynolds and Alexander Melnikov and Natalia Kliushkina and Carlos Araya and Siqi Yan and Orion Reblitz-Richardson},
  journal={arXiv preprint arXiv:2009.07896},
  year={2020}
}

@article{izbicki2017converting,
  title={{Converting High-Dimensional Regression to High-Dimensional Conditional Density Estimation}},
  author={Izbicki, Rafael and Lee, Ann B},
  journal={Electronic Journal of Statistics},
  volume={11},
  pages={2800--2831},
  year={2017}
}

@Manual{R,
  title = {{\proglang{R}: {A} Language and Environment for Statistical Computing}},
  author = {{\proglang{R} Core Team}},
  organization = {\proglang{R} Foundation for Statistical Computing},
  address = {Vienna, Austria},
  year = {2017},
  url = {https://www.R-project.org/},
}

@article{shapley1953value,
  title={{A Value for N-Person Games}},
  author={Shapley, Lloyd S.},
  journal={Contributions to the Theory of Games},
  volume={2},
  number={28},
  pages={307--317},
  year={1953}
}

@Manual{Parsnip,
    title = {{\pkg{parsnip}: A Common API to Modeling and Analysis Functions}},
    author = {Max Kuhn and Davis Vaughan},
    year = {2024},
    note = {\proglang{R} package version 1.3.1},
    url = {https://CRAN.R-project.org/package=parsnip},
  }

@article{aas2019explaining,
  title={{Explaining Individual Predictions when Features are Dependent: More Accurate Approximations to Shapley Values}},
  author={Aas, Kjersti and Jullum, Martin and L{\o}land, Anders},
  journal={{Artificial Intelligence}},
  volume={298},
  year={2021}
}

@article{mitchell2022sampling,
  title={{Sampling Permutations for Shapley Value Estimation}},
  author={Mitchell, Rory and Cooper, Joshua and Frank, Eibe and Holmes, Geoffrey},
  journal={Journal of Machine Learning Research},
  volume={23},
  number={43},
  pages={1--46},
  year={2022}
}

@article{chen2023algorithms,
  title={{Algorithms to Estimate Shapley Value Feature Attributions}},
  author={Chen, Hugh and Covert, Ian C and Lundberg, Scott M and Lee, Su-In},
  journal={Nature Machine Intelligence},
  volume={5},
  number={6},
  pages={590--601},
  year={2023},
  publisher={Nature Publishing Group UK London}
}

@book{kroese2013handbook,
  title={Handbook of Monte Carlo Methods},
  author={Kroese, Dirk P and Taimre, Thomas and Botev, Zdravko I},
  year={2013},
  publisher={John Wiley \& Sons}
}

@inproceedings{lundberg2017unified,
  title={A Unified Approach to Interpreting Model Predictions},
  author={Lundberg, Scott M and Lee, Su-In},
  booktitle={Advances in Neural Information Processing Systems},
  pages={4765--4774},
  year={2017}
}

@Manual{xgboost,
  title = {\pkg{xgboost}: Extreme Gradient Boosting},
  author = {Tianqi Chen and Tong He and Michael Benesty and Vadim Khotilovich and Yuan Tang and Hyunsu Cho and Kailong Chen and Rory Mitchell and Ignacio Cano and Tianyi Zhou and Mu Li and Junyuan Xie and Min Lin and Yifeng Geng and Yutian Li and Jiaming Yuan},
  year = {2024},
  note = {\proglang{R} package version 1.7.9.1},
  url = {https://CRAN.R-project.org/package=xgboost},
}

@article{olsen2024comparative,
  title={A Comparative Study of Methods for Estimating Model-Agnostic Shapley Value Explanations},
  author={Olsen, Lars Henry Berge and Glad, Ingrid Kristine and Jullum, Martin and Aas, Kjersti},
  journal={Data Mining and Knowledge Discovery},
  pages={1--48},
  year={2024},
  publisher={Springer}
}

@article{chen2020true,
  title={True to the Model or True to the Data?},
  author={Chen, Hugh and Janizek, Joseph D. and Lundberg, Scott and Lee, Su-In},
  journal={arXiv preprint arXiv:2006.16234},
  year={2020}
}

@inproceedings{williamson2020efficient,
  title={Efficient Nonparametric Statistical Inference on Population Feature Importance Using Shapley Values},
  author={Williamson, Brian and Feng, Jean},
  booktitle={International Conference on Machine Learning},
  pages={10282--10291},
  year={2020},
  organization={PMLR}
}

@inproceedings{frye_shapley-based_2020,
  title={Shapley Explainability on the Data Manifold},
  author={Christopher Frye and Damien de Mijolla and Tom Begley and Laurence Cowton and Megan Stanley and Ilya Feige},
  booktitle={International Conference on Learning Representations},
  year={2021}
}

@article{frye2020asymmetric,
  title={Asymmetric Shapley values: Incorporating Causal Knowledge into Model-Agnostic Explainability},
  author={Frye, Christopher and Rowat, Colin and Feige, Ilya},
  journal={Advances in Neural Information Processing Systems},
  volume={33},
  year={2020}
}

@article{heskes2020causal,
  title={Causal Shapley Values: Exploiting Causal Knowledge to Explain Individual Predictions of Complex Models},
  author={Heskes, Tom and Sijben, Evi and Bucur, Ioan Gabriel and Claassen, Tom},
  journal={Advances in Neural Information Processing Systems},
  volume={33},
  year={2020}
}

@inproceedings{redelmeier2020,
  title={Explaining Predictive Models with Mixed Features Using Shapley Values and Conditional Inference Trees},
  author={Redelmeier, Annabelle and Jullum, Martin and Aas, Kjersti},
  booktitle={International Cross-Domain Conference for Machine Learning and Knowledge Extraction},
  pages={117--137},
  year={2020},
  organization={Springer}
}

@article{HothornCtree,
  author = {Torsten Hothorn and Kurt Hornik and Achim Zeileis},
  title = {Unbiased Recursive Partitioning: A Conditional Inference Framework},
  journal = {Journal of Computational and Graphical Statistics},
  volume = {15},
  number = {3},
  pages = {651-674},
  year  = {2006},
  publisher = {Taylor \& Francis}
}

@article{Olsen2022,
  title={Using Shapley Values and Variational Autoencoders to Explain Predictive Models with Dependent Mixed Features},
  author={Olsen, Lars Henry Berge and Glad, Ingrid Kristine and Jullum, Martin and Aas, Kjersti},
  journal={Journal of Machine Learning Research},
  volume={23},
  number={213},
  pages={1--51},
  year={2022}
}

@inproceedings{Olsen2024improving,
      title={Improving the Weighting Strategy in KernelSHAP}, 
      author={Olsen, Lars Henry Berge and Jullum, Martin},
      booktitle={World Conference on Explainable Artificial Intelligence},
      pages={194--218},
      year={2025},
      organization={Springer}
}

@inproceedings{covert2021improving,
  title={Improving Kernelshap: Practical Shapley Value Estimation Using Linear Regression},
  author={Covert, Ian and Lee, Su-In},
  booktitle={International Conference on Artificial Intelligence and Statistics},
  pages={3457--3465},
  year={2021},
  organization={PMLR}
}

@inproceedings{ivanov_variational_2018,
  title={Variational Autoencoder with Arbitrary Conditioning},
  author={Oleg Ivanov and Michael Figurnov and Dmitry Vetrov},
  booktitle={International Conference on Learning Representations},
  year={2019}
}

@inproceedings{pmlr-v32-rezende14,
  title={Stochastic Backpropagation and Approximate Inference in Deep Generative Models},
  author={Rezende, Danilo Jimenez and Mohamed, Shakir and Wierstra, Daan},
  booktitle={International conference on machine learning},
  pages={1278--1286},
  year={2014},
  organization={PMLR}
}

@inproceedings{kingma2014autoencoding,
  author = {Kingma, Diederik P. and Welling, Max},
  booktitle = {2nd International Conference on Learning Representations, {ICLR} 2014, Banff, AB, Canada, April 14-16, 2014, Conference Track Proceedings},
  title = {{Auto-Encoding Variational Bayes}},
  year = 2014
}

@article{Kingma2019AnIT,
  title={An Introduction to Variational Autoencoders},
  author = {Kingma, Diederik P. and Welling, Max},
  journal={Found. Trends Mach. Learn.},
  year={2019},
  volume={12},
  pages={307-392}
}

@Article{future,
  author = {Henrik Bengtsson},
  title = {A Unifying Framework for Parallel and Distributed Processing in \proglang{R} using Futures},
  year = {2021},
  journal = {The \proglang{R} Journal},
  doi = {10.32614/RJ-2021-048},
  pages = {208--227},
  volume = {13},
  number = {2},
}

@inproceedings{jullum2021groupshapley,
  author = {Jullum, Martin and Redelmeier, Annabelle and Aas, Kjersti},
  booktitle = {Italian Workshop on Explainable Artificial Intelligence 2021},
  editor = {Cataldo Musto and Riccardo Guidotti and Anna Monreale and Giovanni Semeraro},
  organization = {XAI.it},
  pages = {28--43},
  title = {Efficient and Simple Prediction Explanations with GroupShapley: A practical Perspective},
  year = {2021}
}

@incollection{charnes1988extremal,
  title={Extremal Principle Solutions of Games in Characteristic Function Form: Core, Chebychev and Shapley Value Generalizations},
  author={Charnes, A and Golany, B and Keane, M and Rousseau, J},
  booktitle={Econometrics of planning and efficiency},
  pages={123--133},
  year={1988},
  publisher={Springer}
}

@Article{DALEX_R,
    title = {\pkg{DALEX}: Explainers for Complex Predictive Models in \proglang{R}},
    author = {Przemyslaw Biecek},
    journal = {Journal of Machine Learning Research},
    year = {2018},
    volume = {19},
    pages = {1-5},
    number = {84}
}

@Article{iml,
    author = {Christoph Molnar and Bernd Bischl and Giuseppe Casalicchio},
    title = {\pkg{iml}: An \proglang{R} package for Interpretable Machine Learning},
    year = {2018},
    publisher = {Journal of Open Source Software},
    volume = {3},
    number = {26},
    pages = {786},
    journal = {Journal of Open Source Software},
}

@Manual{shapper,
    title = {\pkg{shapper}: Wrapper of \proglang{Python} Library '\pkg{shap}'},
    author = {Szymon Maksymiuk and Alicja Gosiewska and Przemyslaw Biecek},
    year = {2020},
    note = {\proglang{R} package version 0.1.3},
    url = {https://CRAN.R-project.org/package=shapper},
}

@Manual{fastshap,
    title = {\pkg{fastshap}: Fast Approximate Shapley Values},
    author = {Brandon Greenwell},
    year = {2024},
    note = {\proglang{R} package version 0.1.1},
    url = {https://CRAN.R-project.org/package=fastshap},
}

@Manual{kernelshap,
    title = {\pkg{kernelshap}: Kernel SHAP},
    author = {Michael Mayer and David Watson},
    year = {2024},
    note = {\proglang{R} package version 0.7.0},
    url = {https://CRAN.R-project.org/package=kernelshap},
}

@Manual{treeshap,
    title = {\pkg{treeshap}: Compute SHAP Values for Your Tree-Based Models Using the 'TreeSHAP' Algorithm},
    author = {Konrad Komisarczyk and Pawel Kozminski and Szymon Maksymiuk and Przemyslaw Biecek},
    year = {2024},
    note = {\proglang{R} package version 0.3.1},
    url = {https://CRAN.R-project.org/package=treeshap},
}

@Manual{shapviz,
    title = {\pkg{shapviz}: SHAP Visualizations},
    author = {Michael Mayer},
    year = {2024},
    note = {\proglang{R} package version 0.9.6},
    url = {https://CRAN.R-project.org/package=shapviz},
}

@Article{modelStudio,
    author = {Hubert Baniecki and Przemyslaw Biecek},
    title = {\pkg{modelStudio}: Interactive Studio with Explanations for {ML} Predictive Models},
    year = {2019},
    month = {Nov},
    volume = {4},
    number = {43},
    pages = {1798},
    publisher = {The Open Journal},
    journal = {Journal of Open Source Software},
  }

@inproceedings{muschalik2024shapiq,
  title     = {\pkg{shapiq}: Shapley Interactions for Machine Learning},
  author    = {Maximilian Muschalik and Hubert Baniecki and Fabian Fumagalli and
               Patrick Kolpaczki and Barbara Hammer and Eyke H\"{u}llermeier},
  booktitle = {Advances in Neural Information Processing Systems},
  volume={37},
  pages={130324--130357},
  year={2024}
}

@Misc{iBreakDown,
    title = {Do Not Trust Additive Explanations},
    author = {Alicja Gosiewska and Przemyslaw Biecek},
    year = {2019},
    eprint = {arXiv:1903.11420},
    url = {https://arxiv.org/abs/1903.11420},
  }

@Manual{shap,
    title = {\pkg{shap}},
    author = {Scott M Lundberg and Su-In Lee},
    year = {2024},
    note = {\proglang{Python} package version 0.46.0},
    url = {https://github.com/shap/shap},
}

@article{lundberg2020local2global,
  title={From local Explanations to Global Understanding with Explainable AI for Trees},
  author={Lundberg, Scott M. and Erion, Gabriel and Chen, Hugh and DeGrave, Alex and Prutkin, Jordan M. and Nair, Bala and Katz, Ronit and Himmelfarb, Jonathan and Bansal, Nisha and Lee, Su-In},
  journal={Nature Machine Intelligence},
  volume={2},
  number={1},
  pages={2522-5839},
  year={2020},
  publisher={Nature Publishing Group}
}

@Book{ExplanatoryModelAnalysis,
  author = {Przemyslaw Biecek and Tomasz Burzykowski},
  title = {{Explanatory Model Analysis}},
  publisher = {Chapman and Hall/CRC, New York},
  year = {2021},
  isbn = {9780367135591},
  url = {https://pbiecek.github.io/ema/},
}

@article{strumbelj2010efficient,
  title={An Efficient Explanation of Individual Classifications Using Game Theory},
  author={{\v{S}}trumbelj, Erik and Kononenko, Igor},
  journal={The Journal of Machine Learning Research},
  volume={11},
  pages={1--18},
  year={2010},
  publisher={JMLR. org}
}

@article{vstrumbelj2014explaining,
  title={Explaining Prediction Models and Individual Predictions with Feature Contributions},
  author={{\v{S}}trumbelj, Erik and Kononenko, Igor},
  journal={Knowledge and Information Systems},
  volume={41},
  pages={647--665},
  year={2014},
  publisher={Springer}
}

@inproceedings{shrikumar2017learning,
  title={Learning Important Features Through Propagating Activation Differences},
  author={Shrikumar, Avanti and Greenside, Peyton and Kundaje, Anshul},
  booktitle={International conference on machine learning},
  pages={3145--3153},
  year={2017},
  organization={PMlR}
}

@article{pearl1995causal,
  title={Causal Diagrams for Empirical Research},
  author={Pearl, Judea},
  journal={Biometrika},
  volume={82},
  number={4},
  pages={669--688},
  year={1995},
  publisher={Oxford University Press}
}

@inproceedings{pearl2012calculus,
  title={The Do-Calculus Revisited},
  author={Pearl, Judea},
  booktitle={Proceedings of the Twenty-Eighth Conference on Uncertainty in Artificial Intelligence},
  pages={3--11},
  year={2012}
}

@article{redell2019shapley,
  title={Shapley Decomposition of R-squared in Machine Learning Models},
  author={Redell, Nickalus},
  journal={arXiv preprint arXiv:1908.09718},
  year={2019}
}

@Manual{gautier2023rpy2,
  author = {Laurent Gautier},
  title = {\pkg{rpy2}: A \proglang{Python} interface to \proglang{R}},
  year = {2024},
  note = {\proglang{Python} library Version 3.5.17},
  url= {https://rpy2.github.io/},
}

@article{au2022grouped,
  title={Grouped Feature Importance and Combined Features Effect Plot},
  author={Au, Quay and Herbinger, Julia and Stachl, Clemens and Bischl, Bernd and Casalicchio, Giuseppe},
  journal={Data Mining and Knowledge Discovery},
  volume={36},
  number={4},
  pages={1401--1450},
  year={2022},
  publisher={Springer}
}

@inproceedings{goldwasser2024stabilizing,
  title={Stabilizing Estimates of Shapley Values with Control Variates},
  author={Goldwasser, Jeremy and Hooker, Giles},
  booktitle={World Conference on Explainable Artificial Intelligence},
  pages={416--439},
  year={2024},
  organization={Springer}
}

@article{Sellereite2019,  
    year = {2019}, 
    publisher = {The Open Journal}, 
    volume = {5}, 
    number = {46}, 
    pages = {2027}, 
    author = {Nikolai Sellereite and Martin Jullum}, 
    title = {\pkg{shapr}: An \proglang{R}-package for Explaining Machine Learning Models with Dependence-Aware Shapley Values}, 
    journal = {Journal of Open Source Software} 
}

@Manual{patchwork,
    title = {\pkg{patchwork}: The Composer of Plots},
    author = {Thomas Lin Pedersen},
    year = {2024},
    note = {\proglang{R} package version 1.3.0},
    url = {https://CRAN.R-project.org/package=patchwork},
  }

@Article{ranger,
    title = {{ranger}: A Fast Implementation of Random Forests for High Dimensional Data in {C++} and {R}},
    author = {Marvin N. Wright and Andreas Ziegler},
    journal = {Journal of Statistical Software},
    year = {2017},
    volume = {77},
    number = {1},
    pages = {1--17},
  }

@Manual{mgcv,
    title = {\pkg{mgcv}: Mixed GAM Computation Vehicle with Automatic Smoothness Estimation},
    author = {Simon Wood},
    year = {2025},
    note = {\proglang{R} package version 1.9.1},
    url = {https://CRAN.R-project.org/package=mgcv},
  }

@Manual{workflows,
    title = {\pkg{workflows}: Modeling Workflows},
    author = {Davis Vaughan and Simon Couch},
    year = {2025},
    note = {\proglang{R} package version 1.2.0},
    url = {https://CRAN.R-project.org/package=workflows},
  }

@Manual{tidymodels,
    title = {\pkg{Tidymodels}: a collection of packages for modeling and machine learning using \pkg{tidyverse} principles.},
    author = {Max Kuhn and Hadley Wickham},
    url = {https://www.tidymodels.org},
    year = {2020},
  }

@Article{RJ-2021-048,
    author = {Henrik Bengtsson},
    title = {A Unifying Framework for Parallel and Distributed Processing in \proglang{R} using Futures},
    year = {2021},
    journal = {The \proglang{R} Journal},
    pages = {208--227},
    volume = {13},
    number = {2},
  }

@Manual{progressr,
    title = {\pkg{progressr}: An Inclusive, Unifying API for Progress Updates},
    author = {Henrik Bengtsson},
    year = {2024},
    note = {\proglang{R} package version 0.15.1},
    url = {https://CRAN.R-project.org/package=progressr},
  }

@Manual{cli,
    title = {\pkg{cli}: Helpers for Developing Command Line Interfaces},
    author = {Gábor Csárdi},
    year = {2025},
    note = {R package version 3.6.4},
    url = {https://CRAN.R-project.org/package=cli},
  }

@Manual{pkgdown,
    title = {\pkg{pkgdown}: Make Static HTML Documentation for a Package},
    author = {Hadley Wickham and Jay Hesselberth and Maëlle Salmon and Olivier Roy and Salim Brüggemann},
    year = {2024},
    note = {R package version 2.1.1},
    url = {https://CRAN.R-project.org/package=pkgdown},
  }

@Book{ggplot2,
    author = {Hadley Wickham},
    title = {ggplot2: Elegant Graphics for Data Analysis},
    publisher = {Springer-Verlag New York},
    year = {2016},
    isbn = {978-3-319-24277-4},
    url = {https://ggplot2.tidyverse.org},
  }


\newpage

\begin{appendix}

\section{Additional approaches and functionality}

\subsection[getmodelspecs]{Specification of the \code{get\_model\_specs} argument}
\label{app:get_model_specs}

As described in the main paper, models not natively supported can be explained by supplying a custom prediction function through the \code{predict_model} argument to \code{explain()} or \code{explain_forecast()}. 
The \code{get_model_specs} argument of the same two functions is an optional argument which allows verifying that the input data has the required columns and correct format. 
Such checks are already automatically performed for most natively supported models, and we strongly recommend enabling these checks also for custom predictive models.
The \code{get_model_specs} function takes the following arguments:
\begin{itemize}
    \item \code{labels}: A vector with the feature names to compute Shapley values for.
    \item \code{classes}: A named vector with the labels as names and the class type as elements.
\item \code{factor_levels}: A named list with the labels as names and vectors with the factor levels as elements (NULL for any numeric features).
\end{itemize}
Omitting \code{get_model_specs} for custom models, disables feature checks.
A message will be shown to indicate this, unless \code{verbose = NULL}.

\subsection[predictmodel for forecasting]{Specification of the \code{predict\_model} argument for forecasting}
\label{app:predict_model_forecast}

The \code{explain_forecast()} function includes native support for predicting models of class \code{stats::Arima} and \code{stats::ar}.
 These functions, available through \code{shapr:::predict\_model.ar}/\newline \code{shapr:::predict\_model.Arima}, can be used as basis for custom implementations for other model classes. 
The \code{predict\_model} function for \code{explain\_forecast()} takes the following arguments:
\begin{itemize}
    \item \code{x}: The model to be used for prediction.
    \item \code{newdata}: The new lagged data which is to be imputed in place for \code{y}, where the number of lags per variable corresponds to \code{explain\_lags\$y}.
    \item \code{newreg}: The new lagged data which is to be imputed in place for \code{xreg}, where the number of lags per variable corresponds to \code{explain\_lags\$xreg}. This also contains data for exogenous regressors through the forecast horizon.
    \item \code{horizon}: The forecast horizon the function is expected to produce a forecast for.
    \item \code{explain\_idx}: A vector containing $\mathcal{T}_{\text{explain}}$ as provided to \code{explain\_forecast}.
    \item \code{explain\_lags}: A list containing two items, \code{y} and \code{xreg} which are the number of lags to be explained per variable in \code{y} and \code{xreg}, respectively.
    \item \code{y}: The full dataset in \code{y}, allowing models which require the full data as basis for the imputation to be used.
    \item \code{xreg}: The full dataset in \code{xreg}, allowing models which require the full data as basis for the imputation to be used.
\end{itemize}

\end{appendix}


\end{document}